\documentclass[reprint, NumberedRefs]{JASA}

\usepackage{amsmath,amssymb,amsfonts}
\usepackage{algpseudocode}
\usepackage{array}
\usepackage[caption=false,font=footnotesize]{subfig}
\usepackage{url}
\usepackage{stmaryrd}
\usepackage{tcolorbox}
\usepackage{graphicx,color,psfrag,multirow}
\usepackage{subfloat}
\usepackage{url}
\usepackage{hyperref}
\usepackage{amssymb}
\usepackage{dashrule}
\usepackage{natbib} 

\allowdisplaybreaks[4]

\newcommand{\bs}[1]{\boldsymbol{#1}}

\makeatletter
\newcommand{\citeinline}[1]{%
	\textnormal{[%
		\begingroup
		\let\NAT@spacechar\relax   %
		\citenum{#1}%
		\endgroup
		]}%
}
\makeatother

\makeatletter
\def\titleblock@produce{%
	\begingroup
	\ltx@footnote@pop
	\def\@mpfn{mpfootnote}%
	\def\thempfn{\thempfootnote}%
	\c@mpfootnote\z@
	\let\@makefnmark\frontmatter@makefnmark
	\frontmatter@setup
	\thispagestyle{titlepage}\label{FirstPage}%
	\frontmatter@title@produce
	\groupauthors@sw{%
		\frontmatter@author@produce@group
	}{%
		\frontmatter@author@produce@script
	}%
	\frontmatter@RRAPformat{%
		\expandafter\produce@RRAP\expandafter{\@date}%
		\expandafter\produce@RRAP\expandafter{\@received}%
		\expandafter\produce@RRAP\expandafter{\@revised}%
		\expandafter\produce@RRAP\expandafter{\@accepted}%
		\expandafter\produce@RRAP\expandafter{\@published}%
	}%
	\frontmatter@abstract@produce
	\vskip-4pt
	\@ifx@empty\@pacs{}{%
		\@pacs@produce\@pacs
	}%
	\@ifx@empty\@keywords{}{%
		\@keywords@produce\@keywords
	}%
	\par
	\frontmatter@finalspace
	\endgroup
	\pagestyle{reprint}
}
\makeatother

\begin{document}

\title{Hankel-FNO: Fast Underwater Acoustic Charting Via Physics-Encoded Fourier Neural Operator}

\author{Yifan Sun}

\author{Lei Cheng}

\author{Jianlong Li}
\affiliation{College of Information Science and Electronic Engineering, Zhejiang University, Hangzhou, China}

\author{Peter Gerstoft}
\affiliation{Scripps Institution of Oceanography, University of California San Diego, La Jolla, California 92093, USA}


\begin{abstract}
Fast and accurate underwater acoustic charting is crucial for downstream tasks such as environment-aware sensor placement optimization and autonomous vehicle path planning. Conventional methods rely on computationally expensive while accurate numerical solvers, which are not scalable for large-scale or real-time applications. Although deep learning-based surrogate models can accelerate these computations, they often suffer from limitations such as fixed-resolution constraints or dependence on explicit partial differential equation formulations. These issues hinder their applicability and generalization across diverse environments. We propose Hankel-FNO, a Fourier Neural Operator (FNO)-based model for efficient and accurate acoustic charting. By incorporating sound propagation knowledge and bathymetry, our method has high accuracy while maintaining high computational speed. Results demonstrate that Hankel-FNO outperforms traditional solvers in speed and surpasses data-driven alternatives in accuracy, especially in long-range predictions. Experiments show the model’s adaptability to diverse environments and sound source settings with minimal fine-tuning.
\end{abstract}

\maketitle

\section{\label{sec:1} Introduction}

Underwater acoustic maps\cite{zhu2024strategic}, which map sensor locations to sound transmission features (e.g., transmission loss and detection probability) across depths and ranges, are essential for downstream tasks such as environment-aware sensor placement optimization and autonomous vehicle path planning\cite{stevens2023optimally}. Conventional methods generate these maps by repeatedly running computationally expensive numerical solvers, e.g., range-dependent acoustic model (RAM)\cite{collins1995user}, under varying environmental and source configurations\cite{jensen2011computational}. While accurate, these solvers become computationally prohibitive when fine-scale features, such as interference fringes\cite{kinsler2000fundamentals}, need to be resolved. This computational burden grows further when scaling to large ocean regions or enabling real-time applications, making it critical to develop methods that accelerate acoustic charting while preserving precision at finer scales.

Recent advances in deep learning offer promising alternatives due to their fast inference speeds\cite{chen2023towards, raissi2019physics, lu2021learning, li2020fourier, gupta2023towards,li2023transformer}. Two approaches have emerged. The first uses neural networks, such as UNet\cite{ronneberger2015u}, to directly map inputs to outputs\cite{chen2023towards}. While effective for fixed-resolution data, it lacks physical constraints, often producing results that deviate from real-world phenomena, and cannot generalize to arbitrary resolutions.
The second approach addresses these limitations by exploring implicit neural representations\cite{sitzmann2020implicit}, which are resolution-independent. By incorporating the physical laws (partial differential equations (PDEs) and boundary conditions) in the loss function, physics-informed neural networks (PINNs) have been developed\cite{raissi2019physics}. PINNs have shown success in scientific simulations\cite{cuomo2022scientific}, including sound pressure field prediction\cite{yoon2024predicting,duan2024spatial,olivieri2024physics,verburg2024optical}. However, they face challenges: they require explicit knowledge of the governing PDEs, are hard to train, and struggle with high-frequency and multi-scale PDEs\cite{wang2022and}. In the context of acoustic charting, where the desired output is transmission loss (TL), there is no explicit PDE that governs the feature directly. This underscores the need for a more flexible and scalable framework for TL prediction, while maintaining physics consistency and resolution independence.

Neural operators have emerged as a promising solution to these challenges\cite{kovachki2023neural}. Unlike traditional neural networks, they learn mappings between function spaces, enabling predictions at arbitrary resolutions. Inspired by Green’s function methods for solving PDEs\cite{evans2022partial}, neural operators approximate this process using neural networks, bypassing the need for predefined grids or meshes. Among them, the Fourier Neural Operator (FNO)\cite{li2020fourier} is particularly promising and has been applied to seismic waveform modeling\cite{yang2021seismic} and weather forecasting\cite{pathak2022fourcastnet}. FNO operates in the Fourier space, allowing it to efficiently capture long-range dependencies in data, a limitation of traditional neural networks. These capabilities make FNO suited for high-dimensional and complex spatiotemporal problems. Despite its potential, FNO has not yet been explored as a surrogate model for fast acoustic charting. Furthermore, the integration of sound propagation knowledge and bathymetry into the FNO framework remains unaddressed.

We propose Hankel-FNO, a novel framework for fast underwater acoustic charting that incorporates physical knowledge of sound transmission and environmental factors. Specifically, we encode bathymetry into sound speed field (SSF) as one input channel and use a Hankel function encoding as another. The Hankel function encoding, inspired by the parabolic equation methods\cite{jensen2011computational}, captures physical characteristics of sound propagation. Since our focus is on TL charting, only the amplitude of the Hankel function is incorporated into the input. By embedding this additional physical information, Hankel-FNO outperforms purely data-driven operator learning methods while remaining significantly faster than conventional numerical solvers.

To enhance the model's applicability across diverse environments, we explore a fine-tuning strategy\cite{zhang2023learning,wang2024transfer}. Fine-tuning a pretrained model with limited data from a new scenario (e.g., ocean environments or source parameters) demonstrates that Hankel-FNO adapts to varied settings with minimal training effort while maintaining accuracy. Numerical experiments using real-world data demonstrate the good performance of Hankel-FNO. Furthermore, ablation studies confirm that the Hankel function encoding improves accuracy, particularly for long-ranges.

The remainder of this paper is organized as follows. In Sec.~\ref{sec:2}, we introduce the acoustic charting problem and review the fundamentals about conventional (RAM) and neural solver (neural operator). Sec.~\ref{sec:3} presents our physics-informed FNO, detailing the model architecture as well as the training and fine-tuning strategies. In Sec.~\ref{sec:4}, experimental results are demonstrated to showcase the superior accuracy and rapid transferability of our proposed model.

\section{\label{sec:2} Conventional and Neural Solver}
In this section, we first introduce the formulation of acoustic charting (Sec.~\ref{sec:2.1}), and then present an overview of: 1) Parabolic-equation modeling, which embodies mature numerical techniques for wave propagation modeling (Sec.~\ref{sec:2.1}). 2) The FNO, which synergizes Green's function method with Fourier transform, demonstrating promising expressive power for advancing scientific computation (Sec.~\ref{sec:2.2}).

\subsection{\label{sec:2.1} Acoustic Charting and Conventional Solver}
We aim to achieve fast TL charting\cite{zhu2024strategic}, with SSF $c(r,z)$ as the input. The problem is written as:
\begin{align}
	c(r,z)\rightarrow TL(r,z),
\end{align}
where $(r,z)$ denote the range and depth coordinates.
For conventional methods, it first predicts the full sound pressure via repeatedly running the numerical solver, and then calculates the TL. In the following, we derive the standard parabolic equation, and demonstrate the role of Hankel function in classic sound propagation modeling. Further advanced parabolic-equation modeling methods are in Ref.~[\citen{jensen2011computational}].

Typically, for a medium with constant density, a three dimensional (3D) Helmholtz equation in cylindrical coordinates ($r,\varphi,z$) is written as\cite{jensen2011computational}:
\begin{align}
	\frac{1}{r}\frac{\partial}{\partial r} \left (r\frac{\partial p}{\partial r}\right ) + \frac{1}{r^2}\frac{\partial^2 p}{\partial \varphi^2} + \frac{\partial^2 p}{\partial z^2} + \frac{\omega^2}{c^2}p=0,
	\label{eq:3d-helmholtz}
\end{align}
where $p(r,\varphi,z)$ is the acoustic pressure, $c(r,\varphi,z)$ is the sound speed, and $\omega$ is the angular frequency.
Assuming azimuthal symmetry, \eqref{eq:3d-helmholtz} is reformulated as:
\begin{align}
	\frac{\partial^2 p}{\partial r^2}+\frac{1}{r}\frac{\partial p}{\partial r}+\frac{\partial^2 p}{\partial z^2} + \frac{\omega^2}{c^2}p=0.
	\label{eq:2d-helmholtz}
\end{align}

\begin{figure*}[t]
	\center
	\includegraphics[width=1.9\reprintcolumnwidth]{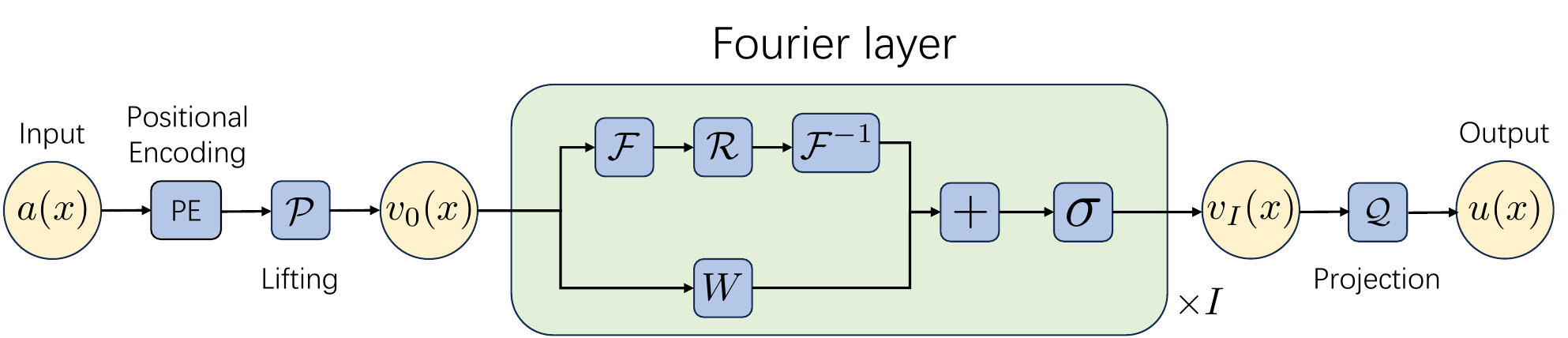}
	\caption{The architecture of Fourier neural operator. The detailed implementation is in Ref.~\citeinline{li2020fourier}.}
	\label{fig:neuraloperator_pipeline}
	\hrule
\end{figure*}

In Ref.~[\citen{jensen2011computational}], it is assumed that the acoustic pressure can be factorized as
\begin{align}
	p(r,z) = \psi(r,z)H_0^{(1)}(k_0r),
	\label{eq:p-phih}
\end{align}
where $\psi(r,z)$ is an envelope function that slowly varies in range and $H_0^{(1)}(k_0r)$ is the Hankel function that satisfies the Bessel differential equation:
\begin{align}
	\frac{\partial^2 H_0^{(1)}(k_0r)}{\partial r^2}+\frac{1}{r}\frac{\partial H_0^{(1)}(k_0r)}{\partial r}+k_0^2H_0^{(1)}(k_0r)=0.
	\label{eq:hankel_bessel}
\end{align}
In the far field ($k_0r\gg1$), the Hankel function is asymptotically:
\begin{align}
	H_0^{(1)}(k_0r)\simeq\sqrt{\frac{2}{\pi k_0r}}e^{i(k_0r-\frac{\pi}{4})}.
	\label{eq:hankel-farfield}
\end{align}
Substituting \eqref{eq:p-phih} into \eqref{eq:2d-helmholtz}, and utilizing the property in \eqref{eq:hankel_bessel} and the far field assumption in \eqref{eq:hankel-farfield}, we obtain
\begin{align}
	\frac{\partial^2 \psi}{\partial r^2}+2ik_0\frac{\partial \psi}{\partial r}+\frac{\partial^2 \psi}{\partial z^2}+k_0^2(n^2-1)\psi=0.
	\label{eq:simplified-elliptic-wave-equation}
\end{align}
Finally, a small-angle approximation is introduced:
\begin{align}
	\frac{\partial^2\psi}{\partial r^2}\ll 2ik_0\frac{\partial \psi}{\partial r}.
	\label{eq:small-angle}
\end{align}
Based on \eqref{eq:small-angle}, \eqref{eq:simplified-elliptic-wave-equation} is rewritten as
\begin{align}
	2ik_0\frac{\partial \psi}{\partial r}+\frac{\partial^2 \psi}{\partial z^2}+k_0^2(n^2-1)\psi=0,
\end{align} 
which is called standard parabolic equation and can be solved numerically (e.g., split-step Fourier technique\cite{hardin1973applications}, finite-difference / finite-element methods\cite{li2017numerical}). 

\subsection{\label{sec:2.2} Fourier Neural Operator}
Neural operators\cite{li2020fourier,kovachki2023neural,anandkumar2020neural} have emerged as a promising approach for solving complex physical systems. First, built on a data-driven framework, they avoid domain-specific approximations (e.g., small-angle approximation in \eqref{eq:small-angle}), making them general and versatile across different scenarios. Second, after training, neural operators require only a single forward pass to perform inference, achieving significant speedups compared to conventional numerical solvers.\cite{azizzadenesheli2024neural}

The goal of neural operator is to learn a mapping between two function spaces using finite observations\cite{kovachki2023neural}. 
Let $\mathcal{A}$ and $\mathcal{U}$ denote Banach spaces of functions defined on bounded domains $D\subset\mathbb{R}^{d}$, $D'\subset\mathbb{R}^{d'}$, respectively. The underlying operator $\mathcal{G}^\dagger : \mathcal{A} \to \mathcal{U}$ maps input functions $a\in \mathcal{A}$ to output function $u\in\mathcal{U}$.
The neural operator approximates this mapping through a parametric model
\begin{gather}
	\mathcal{G}_{\bs{\theta}}:\mathcal{A}\to\mathcal{U}, \quad \bs{\theta}\in\mathbb{R}^p,
	\label{eq:a2umapping}
\end{gather}
where $\bs{\theta}$ denotes the learnable parameters of the neural operator. An optimal parameter $\bs{\theta}^\dagger$ is obtained through training such that $\mathcal{G}_{\bs{\theta}^\dagger}\approx\mathcal{G}^\dagger$.

For a PDE-related case, the problem is generally formulated as
\begin{gather}
	(\mathcal{L}_\gamma u)(x)=f(x), \quad x\in D,\nonumber\\
	(\mathcal{B}_{\gamma^{'}} u)(x)=g(x), \quad x\in \partial D,
	\label{eq:general-pde}
\end{gather}
where $\mathcal{L}_\gamma$ and $\mathcal{B}_{\gamma^{'}}$ are differential operators depending on the parameter $\gamma$ and $\gamma^{'}$, respectively, $f(x)$ and $g(x)$ are some fixed functions, and we aim to map $f(x)$ to the solution $u(x)$.
In conventional solvers, boundary conditions are explicitly imposed when solving the PDE. In contrast, neural operator-based methods do not enforce boundary conditions directly. Instead, they learn them implicitly from data.
Specifically, to solve \eqref{eq:general-pde}, a Green's function is first defined as a unique solution to the problem
\begin{gather}
	\mathcal{L}_\gamma G_\gamma(x,\cdot)=\delta_x,
	\label{eq:greens-function}
\end{gather}
where $\delta_x$ denote the delta measure on $\mathbb{R}^d$ centered at the point $x$ and $G_\gamma (x,\cdot)$ denotes that the Green's function depends on the parameter $\gamma$. Based on \eqref{eq:greens-function}, the solution to \eqref{eq:general-pde} is written as
\begin{gather}
	u(x)=\int_DG_\gamma (x,y)f(y)\text{d}y.
	\label{eq:integral-green}
\end{gather}

In complex scenarios, obtaining a closed-form solution for the Green's function is often intractable. 
Consequently, in neural operator learning, a parametric kernel function $\kappa_\phi$ is employed to approximate the solution (Ref.~[\citen{anandkumar2020neural}], Eq. (6)):
\begin{gather}
	u(x)=\int_D\kappa_\phi(x,y)f(y)\text{d}y.
\end{gather}
Within this framework, the neural operator is formulated to learn the mapping from a fixed function $f(x)$ to the solution $u(x)$. 
However, in practice, its architecture is flexible enough to learn mappings between any PDE-relevant inputs and outputs. For instance, it can learn mappings from the parameter $\gamma$ governing the Green’s function in \eqref{eq:integral-green}, or boundary condition data $g$ specified on $\partial D$\cite{li2020fourier,kovachki2023neural}, to a variety of PDE-related outputs, such as the real or imaginary part of the solution\cite{zhang2023learning}, or the residual between the input and the solution\cite{zong2023born}. \textit{We denote all such input data as $a(x)$, and the associated outputs as $u(x)$}.

To capture spatial relationships, positional encoding $\text{PE}$ is first added to $a(x)$ as an additional channel: 
\begin{align}
	a'(x) = \text{PE}(a(x)),
\end{align}
which augments each location with normalized range and depth coordinates, enabling the model to distinguish spatial positions in the input.
Then, in order to effectively extract more informative features, $a'(x)$ is mapped into a higher dimensional space through a learnable lifting operator $\mathcal{P}$ (Ref.~[\citen{anandkumar2020neural}], Eq. (7)):
\begin{align}
	v(x)=\mathcal{P}\left(a'(x)\right).
\end{align}
This lifting operator is implemented using a $1 \times 1$ convolution applied at each spatial location, which increases the channel dimension while preserving spatial resolution.
Furthermore, inspired by state-of-the-art neural network design methodologies, an $I$ layers iterative architecture is used to improve the expressive power (Ref.~[\citen{anandkumar2020neural}], Eq. (8)):
\begin{gather}
	v_{i+1}(x)=\sigma\left(\mathcal{W}(v_i(x))+\int_D\kappa_\phi^{(i)}(x,y)v_i(y)\text{d}y\right),\nonumber\\
	i=0,...,I-1,
	\label{eq:general-neural-operator}
\end{gather}
where $\sigma$ is a non-linear activate function and $\mathcal{W}(\cdot)$ is a linear transform. In the end, $v_I(x)$ is mapped back to the desired space through a learnable projection operator $\mathcal{Q}$ (Ref.~[\citen{anandkumar2020neural}], Eq. (9)):
\begin{align}
	u(x)=\mathcal{Q}(v_I(x)).
\end{align}
Similar to the lifting operator, the projection is also realized by a $1\times 1$ convolution that compresses the feature channels into the output channels.
\begin{figure*}[t]
	\center
	\includegraphics[width=1.9\reprintcolumnwidth]{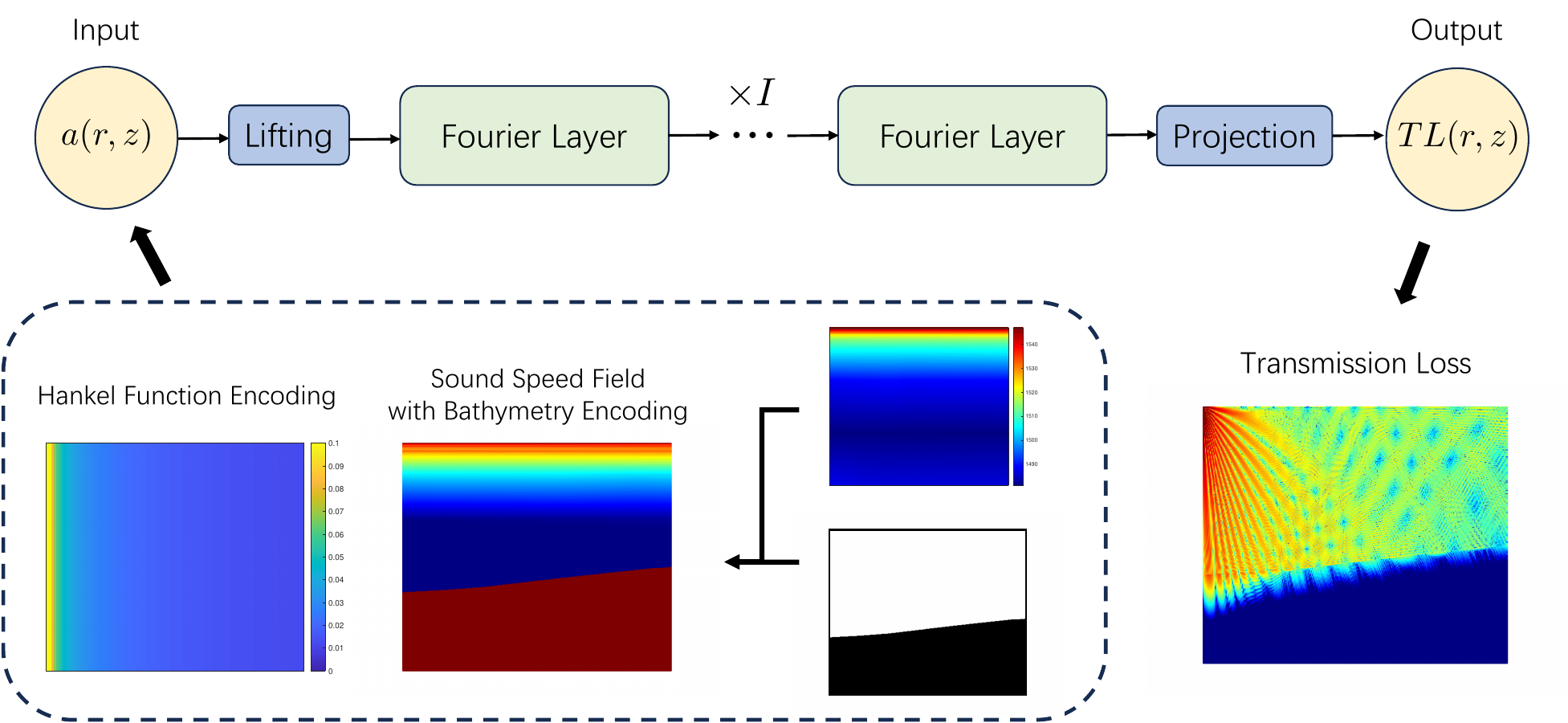}
	\caption{The architecture of Hankel-FNO, where FNO serves as the backbone. Hankel function encoding and SSFs with bathymetry encoding are the input, and transmission losses are the output. In the context of acoustic charting, we mainly consider the variation in transmission losses with respect to range $r$ and depth $z$, so the spatial coordinate is defined as $x=(r,z)$.}
	\label{fig:hankel_pipeline}
	\hrule
\end{figure*}

The main idea of FNO is to solve \eqref{eq:general-neural-operator} in the Fourier domain. By assuming that $\kappa_\phi$ is translation-invariant $\kappa_\phi^{(i)}(x,y)=\kappa_\phi^{(i)}(x-y)$, the integration is transformed into a convolution, which is efficiently computed in the Fourier domain using the fast Fourier transform (FFT) (Ref.~[\citen{li2020fourier}], Eq. (4)):
\begin{align}
	\int_D\kappa_\phi^{(i)}(x-y)v_i(y)\text{d}y&=\mathcal{F}^{-1}(\mathcal{F}(\kappa_\phi^{(i)})\mathcal{F}(v_i))(x) \nonumber\\
	&=\mathcal{F}^{-1}(\mathcal{R}_i\cdot\mathcal{F}(v_i))(x),
	\label{eq:fno-fft}
\end{align}
where $\mathcal{F}$ and $\mathcal{F}^{-1}$ denotes the Fourier transform and its inverse, respectively, and $\mathcal{R}_i$ denotes the Fourier transform of $\kappa_\phi^{(i)}$.
Following the original FNO design\cite{li2020fourier}, only the lowest $k_{\max}$ Fourier modes are retained, with higher frequency modes truncated. This truncation acts as a spectral filter that limits the model to the dominant low-frequency features and reduces overfitting to noise.
Substituting \eqref{eq:fno-fft} into \eqref{eq:general-neural-operator}, we obtain the expression of Fourier layer:
\begin{gather}
	v_{i+1}(x)=\sigma\left(\mathcal{W}v_i(x)+\mathcal{F}^{-1}(\mathcal{R}_i\cdot\mathcal{F}(v_i))(x)\right).
\end{gather}

In conclusion, the procedure of FNO can be summarized as (see Fig.~\ref{fig:neuraloperator_pipeline}):
\begin{enumerate}
	\item Positional encoding: $a'(x)=\text{PE}(a(x))$.
	\item Lifting: $v_0(x)=\mathcal{P}(a'(x))$.
	\item Iterative calculation in Fourier domain:\\
	$v_{i+1}(x)=\sigma\left(\mathcal{W}v_i(x)+\mathcal{F}^{-1}(\mathcal{R}_i\cdot\mathcal{F}(v_i))(x)\right),$\\
	$i=0,...,I-1$.
	\item Projection: $u(x)=\mathcal{Q}(v_I(x))$.
\end{enumerate}

\textit{To effectively apply FNO in a specific scenario, it is crucial to carefully design both the inputs and outputs.} The inputs can incorporate prior knowledge and environmental information, such as source configurations\cite{zhang2023learning} or coarse simulation results\cite{zong2023born}. The output should be tailored to specific objectives through appropriate processing\cite{zhang2023learning}.

\section{\label{sec:3} Physics-Encoded Fourier Neural Operator}
\subsection{\label{sec:3.1} Hankel FNO Architecture}
The proposed model is illustrated in Fig.~\ref{fig:hankel_pipeline}, where FNO serves as the backbone. 
In the context of acoustic charting, we mainly consider the variation in transmission losses with respect to range $r$ and depth $z$, so \textit{the spatial coordinate $x$ is represented as $x=(r,z)$. }
For input-output design, we take SSF $c(r,z)$ with bathymetry encoding and Hankel function encoding as the input $a(r,z)$ and the output $u(r,z)$ is the corresponding transmission loss $TL(r,z)$. 
Detailed analysis of input-output design is as follows.

\subsubsection{\label{sec:3.1.1} Input and Output Design}
A straightforward approach for acoustic charting with FNO is to directly set SSF $c(r,z)$ as input. However, since the pattern of TL is complicated and influenced by multiple environmental factors, relying solely on SSF will lead to low-quality results. 
We enrich the input with additional physical and environmental information, thereby improving the accuracy of the model.

\vspace{0.2cm}
\noindent\textbf{Hankel Function Encoding}

Standard FNO implementations typically augment inputs with positional encoding to capture spatial information. Motivated by this, but aiming to further incorporate physical knowledge, we propose an encoding scheme that embeds the asymptotic form of the Hankel function into the input. This design integrates domain knowledge of sound propagation, enhancing the model’s physical consistency while maintaining computational efficiency.

It is noting that the pressure can be factorized into the product of an envelope function and a Hankel function, see Eq.~\eqref{eq:p-phih}, where the latter has an asymptotic form in the far field, see Eq.~\eqref{eq:hankel-farfield}. This formulation efficiently captures the essential characteristics of sound propagation while maintaining low computational complexity, serving as an ideal choice for encoding. Nevertheless, the phase of the Hankel function undergoes rapid changes with range, thereby increasing the difficulty in training a neural network\cite{yoon2024predicting}. Moreover, since TL is solely dependent on the amplitude of the pressure, the contribution of the phase component is insignificant. Consequently, we incorporate only the amplitude of the Hankel function:
\begin{align}
	\vert H_0^{(1)}(k_0r)\vert=\sqrt{\frac{2}{\pi k_0r}},
	\label{eq:hfe}
\end{align}
which encodes enough sound propagation knowledge while also enhancing training stability.

In practical implementation, once the source frequency is determined, the Hankel function can be computed. Since the Hankel function depends only on the range, we stack it along the depth dimension to create a matrix matching the dimensions of the SSF. Let $\textbf{r}=[r_1, ...,r_N]^\text{T}$ be the range vector, and $\textbf{z}=[z_1, ..., z_M]^\text{T}$ be the depth vector. The resulting stacked matrix $\mathbf{E}_\text{hf}\in \mathbb{R}^{M\times N}$ is constructed as:
\begin{gather}
	\mathbf{E}_\text{hf} = \sqrt{\frac{2}{\pi k_0}}\cdot \bm{1}~[\frac{1}{\sqrt{r_1}} \cdots \frac{1}{\sqrt{r_N}}],
\end{gather}
where $\bm{1}\in \mathbb{R}^M$ is the all-one vector, and the $\mathbf{E}_\text{hf}$ is incorporated as an additional input channel.

\vspace{0.2cm}
\noindent\textbf{Bathymetry Encoding}

Since the sound speed undergoes drastic variations near the sediment layer, deep learning-based methods often struggle due to the spectral bias\cite{rahaman2019spectral}. To mitigate this issue, explicitly incorporating bathymetry information into the algorithm is beneficial. While it can be encoded as an additional channel (as described earlier), directly integrating it with SSF data offers greater convenience. Specifically, for depths beyond the bathymetry, the sound speed is adjusted to that of the sediment layer. This encoding strategy can be formulated as:
$$\mathbf{E}_\text{bty}(r,z)\!=\!
\begin{cases} 
	c(r,z), & z\le D_\text{bty}(r), \\ 
	v_{sed}, & z>D_\text{bty}(r),
\end{cases}
$$
where $D_\text{bty}(r)$ denotes the bathymetry at range $r$, and $v_{sed}$ is the sound speed in the sediment layer, which defaults to 1700 m/s in our experiments.

In conclusion, the input of Hankel-FNO consists of two channels: Hankel function encoding and SSF data with bathymetry information, which can be written as
\begin{align}
	a(r,z)=\{\mathbf{E}_\text{hf}(r,z), \mathbf{E}_\text{bty}(r,z)\}.
	\label{eq:input-a}
\end{align}

\begin{table}[]
	\begin{tcolorbox}[colback=white, colframe=white, boxrule=0pt, arc=0pt, left=0pt, right=0pt, top=0pt, bottom=0pt]
		{\textbf{\color{black} Algorithm 1 \\ Hankel-FNO: Fast Underwater Acoustic Charting with Physics-Encoded Fourier Neural Operator}}
		\\
		\renewcommand{\arraystretch}{0.9}
		\begin{ruledtabular}
			\begin{tabular}{c} 
				\leftline{\textbf {Input:} $c(r,z)$}\\
				\leftline{ // Physics Encoding }\\
				\leftline{$\mathbf{E}_\text{hf}(r,z) = \sqrt{\frac{2}{\pi k_0}} \cdot \bm{1} \left[\frac{1}{\sqrt{r_1}}, \dots, \frac{1}{\sqrt{r_N}} \right]$}\\
				\leftline{$\mathbf{E}_\text{bty}(r,z) = 
					\begin{cases}
						c(r,z), & z \le D_\text{bty}(r) \\
						v_{sed}, & z > D_\text{bty}(r)
					\end{cases}$}\\
				\vspace{0.2cm}
				\leftline{$a(r,z)=\{\mathbf{E}_\text{hf}(r,z), \mathbf{E}_\text{bty}(r,z)\}$.}\\
				\leftline{ // FNO-based Processing }\\
				\leftline{$a'(r,z)=\text{PE}(a(r,z))$}\\
				\leftline{$v_0(r,z)=\mathcal{P}(a'(r,z))$}\\
				\leftline{$v_{i+1}(r,z)=\sigma\left(\mathcal{W}v_i(r,z)+\mathcal{F}^{-1}(\mathcal{R}_i\cdot\mathcal{F}(v_i))(r,z)\right),$}\\
				\leftline{$i=0,...,I-1$}\\
				\leftline{$TL(r,z)=\mathcal{Q}(v_I(r,z))$}
			\end{tabular}
		\end{ruledtabular}
	\end{tcolorbox}
	\label{algorithm1}
\end{table}

\vspace{0.3cm}
\begin{table}[]
	\renewcommand{\arraystretch}{0.9}
	{\textbf{\color{black} Algorithm 2 \\ Implementation Details}}
	\begin{ruledtabular}
		\begin{tabular}{p{8.3cm}}
			\textbf{Positional Encoding}:\\
			Let $\hat{\mathbf{r}}=[\hat{r}_1, ...,\hat{r}_N]^\text{T}\in\mathbb{R}^N$ and $\hat{\mathbf{z}}=[\hat{z}_1, ..., \hat{z}_M]^\text{T}\in\mathbb{R}^M$ \\
			be the normalized coordinate vectors uniformly sampled from the interval $[0, 1]$ along the range and depth dimensions, respectively. We define two positional encoding matrices:
			\begin{gather}
				\textbf{PE}_r=\textbf{1}_M\cdot\hat{\textbf{r}}^\text{T}\in\mathbb{R}^{M\times N},\\
				\textbf{PE}_z=\hat{\textbf{z}}\cdot\textbf{1}^\text{T}_N\in\mathbb{R}^{M\times N},
			\end{gather}
			where $\mathbf{1}_M \in \mathbb{R}^{M}$ and $\mathbf{1}_N \in \mathbb{R}^{N}$ are all-one vectors. Here, $\mathbf{PE}_z$ encodes depth positions along rows, and $\mathbf{PE}_r$ encodes range positions along columns. These are then stacked:
			\parbox[t]{\linewidth}{
				\begin{equation}
					\begin{aligned}
						a'(r,z) \ &=\ \text{PE}(a(r,z)) \\
						&=\ \{a(r,z),\, \mathbf{PE}_r,\, \mathbf{PE}_z\} \in \mathbb{R}^{4\times M\times N}.
					\end{aligned}
				\end{equation}
			}\\[0.5em]
			\textbf{Lifting and Projection Operator}:\\
			Both operators are implemented using $1\times 1$ convolutions at each spatial location. The lifting operator transforms $a'(r,z)\in\mathbb{R}^{4\times M\times N}$ into $v_0(r,z)\in\mathbb{R}^{C\times M\times N}$, and the projection operator maps $v_I(r,z)\in\mathbb{R}^{C\times M\times N}$ to $TL(r,z)\in\mathbb{R}^{M\times N}$. Here, $C$ denotes the number of channels in the lifted feature space.
		\end{tabular}
	\end{ruledtabular}
\end{table}

\noindent\textbf{Output Design}

In the context of acoustic charting, TL serves as the desired output as it can be directly used in downstream tasks\cite{zhu2024strategic,stevens2023optimally}. Therefore, in our algorithm design, we directly set TL as the output:
\begin{align}
	u(r,z)=TL(r,z).
\end{align}
Our proposed Hankel-FNO algorithm is summarized in \textbf{Algorithm 1}.
Further implementation details are in \textbf{Algorithm 2}.

\begin{figure*}[t]
	\center
	\includegraphics[width=1.9\reprintcolumnwidth]{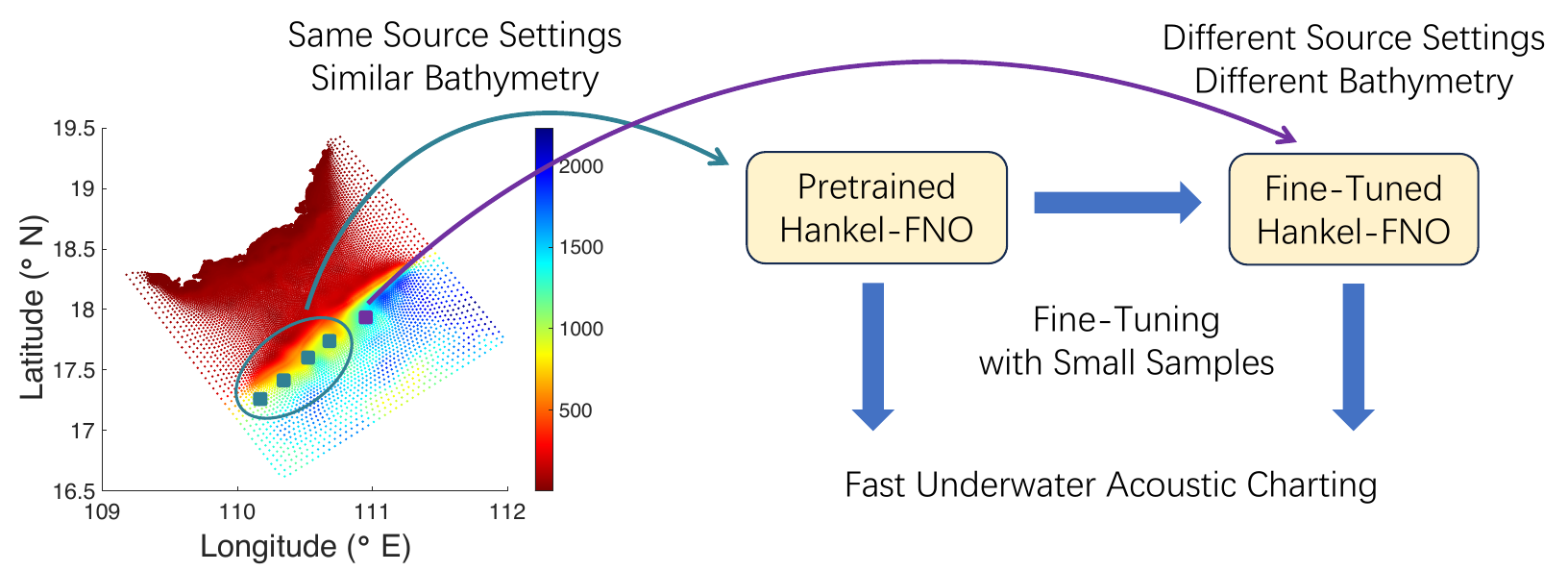}
	\caption{The training and fine-tuning strategy of Hankel-FNO. The model is initially trained on data with same source settings and similar bathymetry. For different scenarios, it is fine-tuned with a small amount of samples while maintaining accuracy. The cyan and purple squares denote the sound source positions of the training and fine-tuning data, respectively.}
	\label{fig:strategy}
	\hrule
\end{figure*}

\subsection{\label{sec:3.2} Training and Fine-tuning Strategy}
Based on \eqref{eq:a2umapping}, given a mapping $\mathcal{G}^\dagger:\mathcal{A}\rightarrow\mathcal{U}$ and $J$ pairs of data $\{a^{(j)},u^{(j)}\}_{j=1}^J$, where $a^{(j)}\in\mathcal{A}$ and $u^{(j)}\in\mathcal{U}$, we aims to find an optimal $\bs{\theta}^{\dagger}$ for the parametric mapping $\mathcal{G}_{\bs{\theta}}$ so that $\mathcal{G}_{\bs{\theta}^\dagger}(a^{(j)})\approx\mathcal{G}^\dagger(a^{(j)})=u^{(j)}$. 
However, although neural operators are theoretically designed to learn mapping between function spaces, their practical implementation necessitates training on finite sets of discrete data points due to computational constraints. In our implementation, we assume $D=D'$ and let $D_S=\{r_s^{(j)},z_s^{(j)}\}_{s=1}^S$ be a $S$-point discretization of domain $D$. Then the discretized input-output pairs are written as $\{a^{(j)}\vert_{D_S},u^{(j)}\vert_{D_S}\}$. The optimization problem is written as:
\begin{align}
	\bs{\theta}^\dagger=\arg\min_{\bs{\theta}}~\sum_{j=1}^J L\left(\mathcal{G}_{\bs{\theta}}\left(a^{(j)}\vert_{D_S}\right),u^{(j)}\vert_{D_S}\right),
\end{align}
where the loss function $L(\cdot,\cdot)$ is defined using $H^1$ Sobolev loss\cite{czarnecki2017sobolev,li2022mno}, which enables better capture of the high frequency details by incorporating higher-order derivatives. Specifically, the $H^1$ Sobolev loss is formulated as\cite{czarnecki2017sobolev}:
\begin{align}
	H^1\!\left(\!\mathcal{G}_{\bs{\theta}}\!\left(\!a^{(j)}\vert_{D_S}\!\right)\!,\! u^{(j)}\vert_{D_S}\!\right)=\sqrt\frac{\mathcal{L}\left(\mathcal{G}_{\bs{\theta}}\left(a^{(j)}\vert_{D_S}\right),u^{(j)}\vert_{D_S}\right)}{\mathcal{N}\left(u^{(j)}\vert_{D_S}\right)},\nonumber
\end{align}
where the loss term $\mathcal{L}(\cdot,\cdot)$ measures the discrepancy between the predicted and true solutions in both function values and partial derivatives up to order $K$, using the Frobenius norm $\|\cdot\|_F$\cite{czarnecki2017sobolev}:
\begin{align}
	\mathcal{L}\!\left(\!\mathcal{G}_{\bs{\theta}}\!\left(a^{(j)}\vert_{D_S}\!\right)\!,\! u^{(j)}\vert_{D_S}\!\right)=\left\|\mathcal{G}_{\bs{\theta}}\left(a^{(j)}\vert_{D_S}\right)-u^{(j)}\vert_{D_S}\right\|^2_F\nonumber\\
	+\sum_{k=1}^K\left\|\mathcal{D}_r^k\mathcal{G}_{\bs{\theta}}\left(a^{(j)}\vert_{D_S}\right)-\mathcal{D}_r^k \left(u^{(j)}\vert_{D_S}\right)\right\|^2_F \nonumber\\
	+\sum_{k=1}^K\left\|\mathcal{D}_z^k\mathcal{G}_{\bs{\theta}}\left(a^{(j)}\vert_{D_S}\right)-\mathcal{D}_z^k \left(u^{(j)}\vert_{D_S}\right)\right\|^2_F,\nonumber
\end{align}
where $\mathcal{D}_r^k$ and $\mathcal{D}_z^k$ denote the $k$-th order derivatives with respect to $r$ and $z$, respectively.
The normalization term $\mathcal{N}(\cdot)$ ensures scale invariance by incorporating the magnitude of the true solution and its derivatives:
\begin{align}
	\mathcal{N}\left(u^{(j)}\vert_{D_S}\right)=\left\|u^{(j)}\vert_{D_S}\right\|^2_F
	+\sum_{k=1}^K\left\|\mathcal{D}_r^k\left(u^{(j)}\vert_{D_S}\right)\right\|^2_F
	\nonumber\\
	+\sum_{k=1}^K\left\|\mathcal{D}_z^k\left(u^{(j)}\vert_{D_S}\right)\right\|^2_F.
	\label{eq:sobolev-loss}
\end{align}
The model parameters $\bs{\theta}$ are optimized through gradient descent methods based on backpropagation.

Theoretically, neural operators learn the mapping from multiple input parameters to their corresponding solutions (as described in Sec.~\ref{sec:2.2}). However, in practice, training such a model requires a large amount of data, which is impractical in ocean acoustics-related scenarios. Furthermore, the ocean environment varies drastically with location and time, and different source configurations (e.g., source depth and frequency) also correspond to distinct propagation patterns. As a result, the collected data exhibit significant distribution differences, increasing the difficulty of model training and leading to instability.
To address these challenges, we adopt a two-stage training and fine-tuning strategy (see Fig.~\ref{fig:strategy}):

\begin{enumerate}
	\item Pretraining for Efficiency: To improve training efficiency and reduce the need for extensive training data, the model is initially trained on data with similar bathymetry, a single source frequency, and a fixed source depth.
	\item Fine-tuning for Adaptability: To enable rapid deployment across different ocean environments and source configurations, we employ a fine-tuning strategy. For a specific scenario, the model is fine-tuned using a limited dataset that encompasses diverse environmental and source conditions. 
	
	Specifically, let $D_{S'}=\{r_s^{(t)},z_s^{(t)}\}_{s=1}^{S'}$ be a ${S'}$-point discretization of the domain $D$. Given a pretrained model $\mathcal{G}_{\bs{\theta}^\dagger}$ and $T$ pairs of fine-tuning data $\{a^{(t)}\vert_{D_{S'}}, u^{(t)}\vert_{D_{S'}}\}_{t=1}^{T}$, the fine-tuning problem is written as:
	\begin{align}
		\hat{\bs{\theta}}=\arg\min_{\bs{\theta}}\sum_{t=1}^T L\left(\mathcal{G}_{\bs{\theta}}\left(a^{(t)}\vert_{D_{S'}}\right),u^{(t)}\vert_{D_{S'}}\right),
	\end{align}
	where the model parameters are updated via gradient descent, and the pretrained parameters $\bs{\theta}^\dagger$ serve as the initialization. In our implementation, all the parameters are fine-tuned. Details of the fine-tuning dataset are in Sec.~\ref{sec:transfer_learning}.
\end{enumerate}

\begin{figure*}[t]
	\center
	\includegraphics[width=2\reprintcolumnwidth]{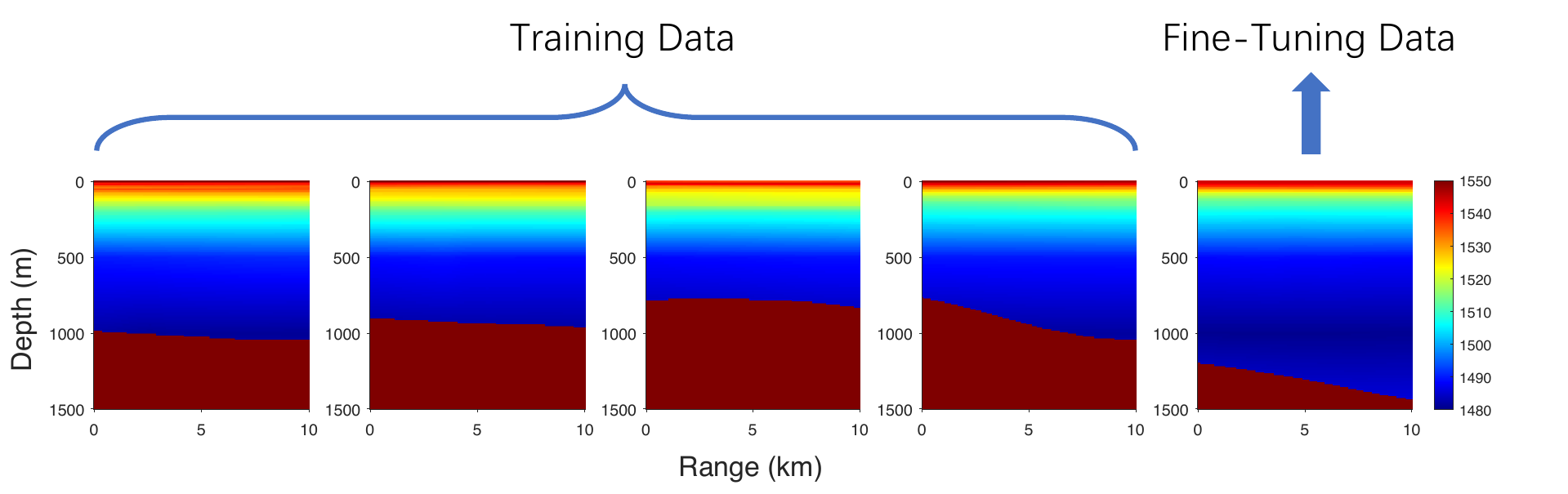}
	\caption{SSF samples along the same bearing from different areas, where the first four are training data with similar bathymetry, and the last one is a fine-tuning sample with deeper bathymetry}
	\label{fig:ssp}
	\hrule
\end{figure*}

\subsection{\label{sec:3.3} Inference}
After training, Hankel-FNO can perform real-time inference through a single forward pass, achieving significant acceleration compared with conventional solvers (see Sec.~\ref{sec:4.2}). As a neural operator framework, it learns the mapping between function spaces and inherently possesses discretization invariance, enabling directly processing input data with arbitrary resolution without requiring architectural modifications or retraining\cite{li2020fourier}. This discretization invariance stems from the architecture design of FNO: as introduced in Sec.~\ref{sec:2.2}, all operations in FNO are either global (Fourier transforms) or element-wise (positional encoding, lifting and projection operators via $1\times 1$ convolutions, and linear transforms), making them independent of the input data dimensions. The only operation that depends on the data size is the frequency-domain filtering in \eqref{eq:fno-fft}, where $\mathcal{R}_i$ operates on Fourier modes. However, due to the mode truncation strategy that retains only the lowest $k_{\max}$ Fourier modes regardless of the input resolution, the frequency-domain data is consistently cropped to the same size, ensuring that the dimensions of $\mathcal{R}_i$ remain fixed. Consequently, given a model $\mathcal{G}_{\bs{\theta^\dagger}}$ trained with low resolution data $\{a^{(j)}\vert_{D_l},u^{(j)}\vert_{D_l}\}$, this capability enables the model to seamlessly perform zero-shot high-resolution inference:
\begin{align}
	u^{(new)}\vert_{D_h}=\mathcal{G}_{\bs{\theta^\dagger}} (a^{(new)}\vert_{D_h}).
\end{align}
where $D_l$ and $D_h$ denote low and high resolution discretization, respectively. 

\section{\label{sec:4} Numerical Results}
\subsection{Experimental Settings}
\subsubsection{\label{sec:4.1.1}3D Environmental Field Data}
\begin{figure}[t]
	\center
	\includegraphics[width=0.80\reprintcolumnwidth]{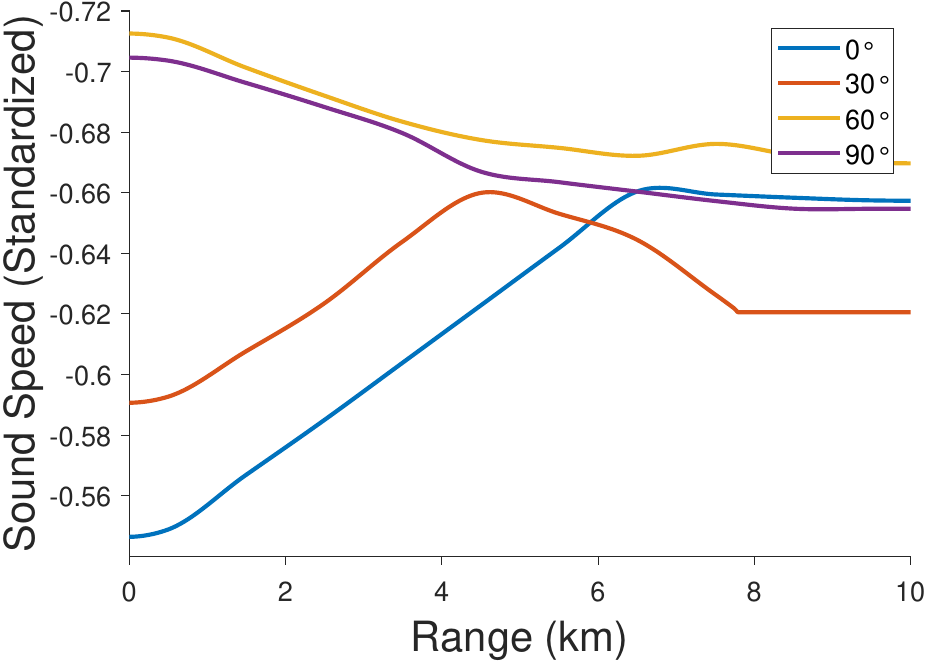}
	\caption{Standardized SSP samples at depth 700 m for four different bearings: $0^{\circ}$, $30^{\circ}$, $60^{\circ}$, and $90^{\circ}$, where $0^{\circ}$ corresponds to the eastward direction.}
	\label{fig:ssp-diffbearings}
	\hrule
\end{figure}

The 3D environmental field data, including temperature and salinity, are from finite volume coastal ocean model\cite{chen2006finite}, while the bathymetry is from the ETOPO1 Global Relief Model\cite{amante2009etopo1} provided by the U.S. National Geophysical Data Center. The data from June 27, 2020 are used in our experiments.
Four areas with similar bathymetry between ($17.2^{\circ}~\text{N}, 110.1^{\circ}~\text{E}$) and ($17.8^{\circ}~\text{N}, 110.7^{\circ}~\text{E}$) are selected for model pretraining, and a fifth area with different bathymetry is used for fine-tuning. Fig.~\ref{fig:ssp} shows the similarities and differences in bathymetry.
To calculate TLs via RAM, environmental field data along $N=36$ bearings at different times are selected. 
Fig.~\ref{fig:ssp-diffbearings} shows the standardized sound speed profile (SSP) in the training dataset at 700 m depth along different bearings.

We prepare two pretraining datasets covering ocean regions of 10 km × 1.5 km and 40 km × 1.5 km, with input-output pair sizes of (200, 150) for the short range and (800, 150) for the long range. The data are extracted from the four cyan positions in Fig.~\ref{fig:strategy}. For each range, the pretraining dataset contains 3456 samples across various time points and bearings. Both datasets are partitioned in a 9:1 ratio, with 3110 samples used to pretrain and 346 data for performance evaluation.
All the data are standardized to zero mean and unit variance to enhance training stability and model convergence. For each input-output pair $\{a^{(j)}\vert_{D_S}, u^{(j)}\vert_{D_S}\}$, the standardized data are
\begin{align}
	\tilde{a}^{(j)}\vert_{D_S}=\frac{a^{(j)}\vert_{D_S}-\mu_a}{\sigma_a},\quad \tilde{u}^{(j)}\vert_{D_S}=\frac{u^{(j)}\vert_{D_S}-\mu_u}{\sigma_u},
\end{align}
where $\mu_a,\sigma_a$ and $\mu_u,\sigma_u$ denote the empirical mean and standard deviation of the inputs and outputs over the entire pretraining dataset, respectively.
The simulation setup of RAM for generating the pretraining data is in Table~\ref{tab:ram_setup}. Details about the dataset and simulation setup for fine-tuning are in Sec.~\ref{sec:transfer_learning}.

\subsubsection{\label{sec:4.1.2}Model Hyperparameters}
The key hyperparameters of Hankel-FNO are listed in Table~\ref{tab:model_hyperparams}. 
The model uses $I=4$ Fourier layers, with a feature dimension of $C=64$. 
The number of retained Fourier modes is $k_{\max}=64$, balancing accuracy and efficiency. 
The model adopts the GELU\cite{dan2016gelu} activation function and the AdamW\cite{loshchilov2017decoupled} optimizer.
Training is performed for 1000 epochs.

\begin{table}[t!]
	\begin{center}
		\caption{The parameter setup of RAM.}
		\begin{tabular}{cc}
			\hline\hline
			Parameter & Value \\
			\hline
			Source depth & 50 m \\
			Source frequency & 200 Hz \\
			Sediment density & 2.0 g/cm$^3$ \\
			Sediment attenuation & 0.8 dB/$\lambda$ \\
			~~~~~~~~Sediment sound speed~~~~~~~~ & ~~~~~~~~1700 m/s~~~~~~~~\\
			\hline\hline				
		\end{tabular}
		\label{tab:ram_setup}
	\end{center}
\end{table}

\begin{table}[t!]
	\begin{center}
		\caption{The hyperparameters of Hankel-FNO.}
		\label{tab:model_hyperparams}
		\begin{tabular}{cc}
			\hline
			Parameter & Value \\
			\hline
			Number of Fourier layers ($I$) & 4 \\
			Feature dimension ($C$) & 64 \\
			Fourier modes ($k_{\max}$) & 64 \\
			Activation function & GELU \\
			Optimizer & AdamW \\
			~~~~~~~~~~~Training epochs~~~~~~~~~~~ & ~~~~~~~~~~~1000~~~~~~~~~~~\\
			\hline\hline
		\end{tabular}
	\end{center}
\end{table}

\subsubsection{Performance Measure}
The results are evaluated using the following three metrics:

\noindent$\blacksquare$ \textbf{Root Mean Square Error (RMSE)}:
\begin{align}
	\text{RMSE}=\sqrt{\frac{1}{Q}\Vert \widehat{TL}-TL\Vert_F^2},
\end{align}
where $TL$ and $\widehat{TL}$ denote the TLs computed via RAM and operator learning methods, respectively, and $Q$ denotes the total number of entries in the TL data.

\vspace{0.2cm}
\noindent$\blacksquare$ \textbf{$H^1$ Sobolev Error}\cite{czarnecki2017sobolev,li2022mno}:
\begin{gather}
	H^1\left(\widehat{TL},TL\right)=\sqrt{\frac{\mathcal{L}\left(\widehat{TL},TL\right)}{\mathcal{N}\left(TL\right)}},\nonumber\\
	\mathcal{L}\left(\widehat{TL},TL\right)=\left\|\widehat{TL}-TL\right\|^2_F
	+\left\|\mathcal{D}_r^1\left(\widehat{TL}\right)-\mathcal{D}_r^1 \left(TL\right)\right\|^2_F \nonumber\\
	+\left\|\mathcal{D}_z^1\left(\widehat{TL}\right)-\mathcal{D}_z^1 \left(TL\right)\right\|^2_F,\nonumber\\
	\mathcal{N}\left(TL\right)=\left\|TL\right\|^2_F
	+\left\|\mathcal{D}_r^1\left(TL\right)\right\|^2_F+\left\|\mathcal{D}_z^1\left(TL\right)\right\|^2_F,
\end{gather}
where $K$ in \eqref{eq:sobolev-loss} is set to 1 to balance the computational cost and accuracy.

\vspace{0.2cm}
\noindent$\blacksquare$ \textbf{Structure Similarity Index Measure (SSIM)}\cite{wang2004image}:
\begin{align}
	SSIM(\widehat{TL},TL)=\frac{(2\mu_{\widehat{TL}}\mu_{TL}+C_1)(2\sigma_{\widehat{TL}TL}+C_2)}{(\mu_{\widehat{TL}}^2+\mu_{TL}^2+C_1)(\sigma_{\widehat{TL}}^2+\sigma_{TL}^2+C_2)}
\end{align}
where $\mu_{\widehat{TL}}$ and $\mu_{TL}$ are the mean of $\widehat{TL}$ and $TL$, respectively, $\sigma_{\widehat{TL}}^2$ and $\sigma_{TL}^2$ are the variance of $\widehat{TL}$ and $TL$, respectively, and $\sigma_{\widehat{TL}TL}$ is the covariance between $\widehat{TL}$ and $TL$. Constant $C_1$ and $C_2$ are used to avoid division by zero. We use the default values in MATLAB, with $C_1=10^{-4}$ and $C_2=9\times10^{-4}$ (see \url{https://www.mathworks.com/help/images/ref/ssim.html}).
All experiments are conducted on a computer with a 5.8 GHz Intel i9 CPU and an NVIDIA GeForce RTX 4090 GPU.

\subsection{\label{sec:4.2} Comparison with Baselines}
In this subsection, we compare the inference time and accuracy between Hankel-FNO and other operator learning methods (OFormer [\citen{li2023transformer}] and vanilla FNO [\citen{li2020fourier}]) at the pretraining stage. Due to the limitation of OFormer to square input data, we train it using data with the size of (150, 150), corresponding to the ocean region of 7.5 km range $\times$ 1.5 km depth. To ensure a fair comparison, we truncate the output of FNO-based methods to (150, 150) before calculating the performance metrics. The comparison of full 10 km TL prediction is demonstrated in Sec.~\ref{sec:4.3}.

\begin{figure}[t]
	\center
	\includegraphics[width=1.0\reprintcolumnwidth]{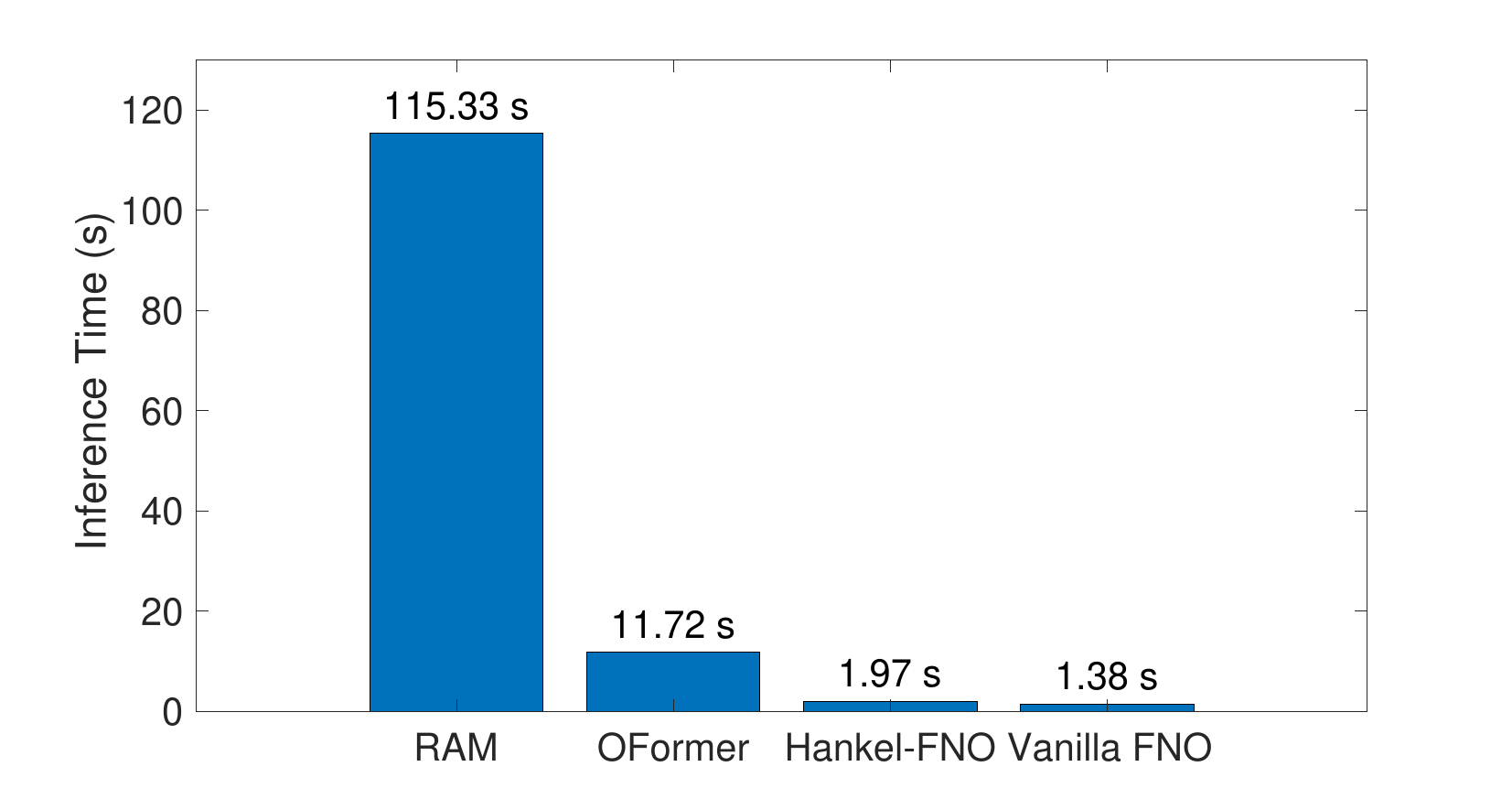}
	\caption{The inference time required of different algorithms to process all input samples $a(r,z)$ and generate their corresponding outputs $u(r,z)$.}
	\label{fig:baseline_inference_time}
	\hrule
\end{figure}

To evaluate the performance of the pretrained model, we use 346 test pairs $\{a(r,z), u(r,z)\}$ from the training region that were not seen in the training stage. The inference time required of different algorithms to process all input samples $a(r,z)$ and generate their corresponding outputs $u(r,z)$ is presented in Fig.~\ref{fig:baseline_inference_time}. The operator learning methods are much faster than the conventional numerical solver. Among them, since FNO-based methods use FFT to efficiently compute the convolution operation, it is faster than OFormer, which is based on the computationally expensive attention mechanisms. Due to the extra encoding, Hankel-FNO is slightly slower than vanilla FNO, but achieves higher accuracy, see subsequent numerical results.

\begin{figure*}[t]
	\center
	\includegraphics[width=2.0\reprintcolumnwidth]{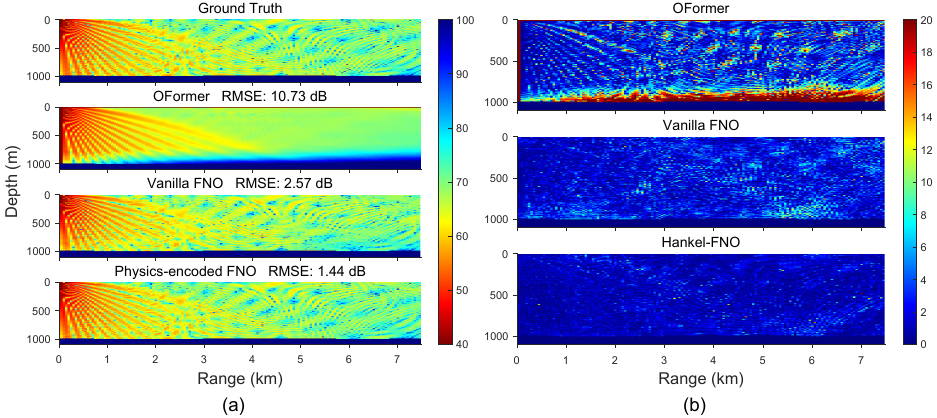}
	\caption{(a) TLs and (b) error surfaces of Hankel-FNO and baseline methods for 7.5 km predictions.}
	\label{fig:baseline_visual_effects}
	\hrule
\end{figure*}

The performance metrics of Hankel-FNO and baseline methods for 7.5 km TL predictions are presented in Table \ref{tab:baseline-10km}, and the generated TLs and error surfaces are in Fig.~\ref{fig:baseline_visual_effects}. The proposed Hankel-FNO outperforms both the vanilla FNO and OFormer across all evaluate metrics. Specifically, OFormer, a Transformer-based model, relies on large datasets to learn data patterns. With limited training data, it struggles to capture the underlying structure, leading to inferior performance. 
In contrast, vanilla FNO surpasses OFormer due to its incorporation of the Green's function, which introduces a PDE-related inductive bias and enhances its modeling capability. 
Hankel-FNO further enhances accuracy by incorporating environmental and sound propagation information, yielding more accurate acoustic transmission modeling.

\begin{table}[t!]
	\begin{center}
		\caption{The performance metrics of Hankel-FNO and baselines for 7.5 km TL predictions.}
		\begin{tabular}{cccc}
			\hline\hline
			  &  $H^1$ ($\times10^{-2}$) ($\downarrow$)~  &  ~RMSE ($\downarrow$)   &  ~~SSIM ($\uparrow$)~~  \\
			\hline
			~~~OFormer~~  &  4.76~  &  ~9.59  &  ~~0.53~~  \\
			\hline
			~~~Vanilla FNO~~  &  2.42~  &  ~2.61  &  ~~0.90~~  \\
			\hline
			~~~Hankel-FNO~~  &  \textbf{0.91}~  &  ~\textbf{1.43}  &  ~~\textbf{0.97}~~  \\
			\hline\hline			
		\end{tabular}
		\label{tab:baseline-10km}
	\end{center}
\end{table}

\begin{table}[t!]
	\begin{center}
		\caption{The performance metrics of FNO with different encodings for 10 km and 40 km TL predictions. $C_\text{bty}$: bathymetry as a separate channel; $E_\text{bty}$: bathymetry encoding; $E_\text{hf}$: Hankel function encoding.}
		\label{tab:ablation-10km-40km}
		\begin{tabular}{cccc}
			\hline\hline
			10 km &  $H^1$ ($\times10^{-2}$) ($\downarrow$)~  &  ~RMSE ($\downarrow$)   &  ~~SSIM ($\uparrow$)~~  \\
			\hline
			~~~w/o Encodings~~  &  2.35~  &  ~2.24  &  ~~0.90~~  \\
			\hline
			~~~w/ $C_\text{bty}$~~ & 1.12~ & ~1.38 & ~~0.97~~ \\
			\hline
			~~~w/ $E_\text{bty}$~~  & 0.91~ & ~1.35 & ~0.97~~\\
			\hline
			~~~w/ $E_\text{bty}$ \& $E_\text{hf}$~~  &  \textbf{0.87}~  &  ~\textbf{1.29}  &  ~~\textbf{0.97}~~  \\
			\hline\hline	
			40 km &  $H^1$ ($\times10^{-1}$) ($\downarrow$)~  &  ~RMSE ($\downarrow$)   &  ~~SSIM ($\uparrow$)~~  \\
			\hline
			~~~w/o Encodings~~  &  4.87~  &  ~4.54  &  ~~0.86~~  \\
			\hline
			~~~w/ $E_\text{bty}$~~  &  3.47~  &  ~4.30  &  ~~0.90~~  \\
			\hline
			~~~w/ $E_\text{bty}$ \& $E_\text{hf}$~~  &  \textbf{0.54}~  &  ~\textbf{4.10}  &  ~~\textbf{0.91}~~  \\
			\hline\hline			
		\end{tabular}
	\end{center}
\end{table}


\begin{figure*}[t]
	\center
	\includegraphics[width=2.0\reprintcolumnwidth]{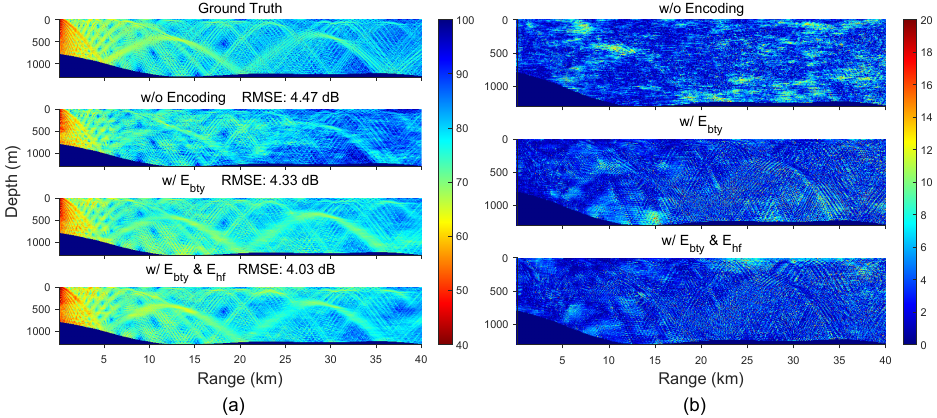}
	\caption{(a) TLs and (b) error surfaces of FNO with different encodings for 40 km predictions.}
	\label{fig:ablation_visual_effects}
	\hrule
\end{figure*}

\begin{figure}[t]
	\center
	\includegraphics[width=1.0\reprintcolumnwidth]{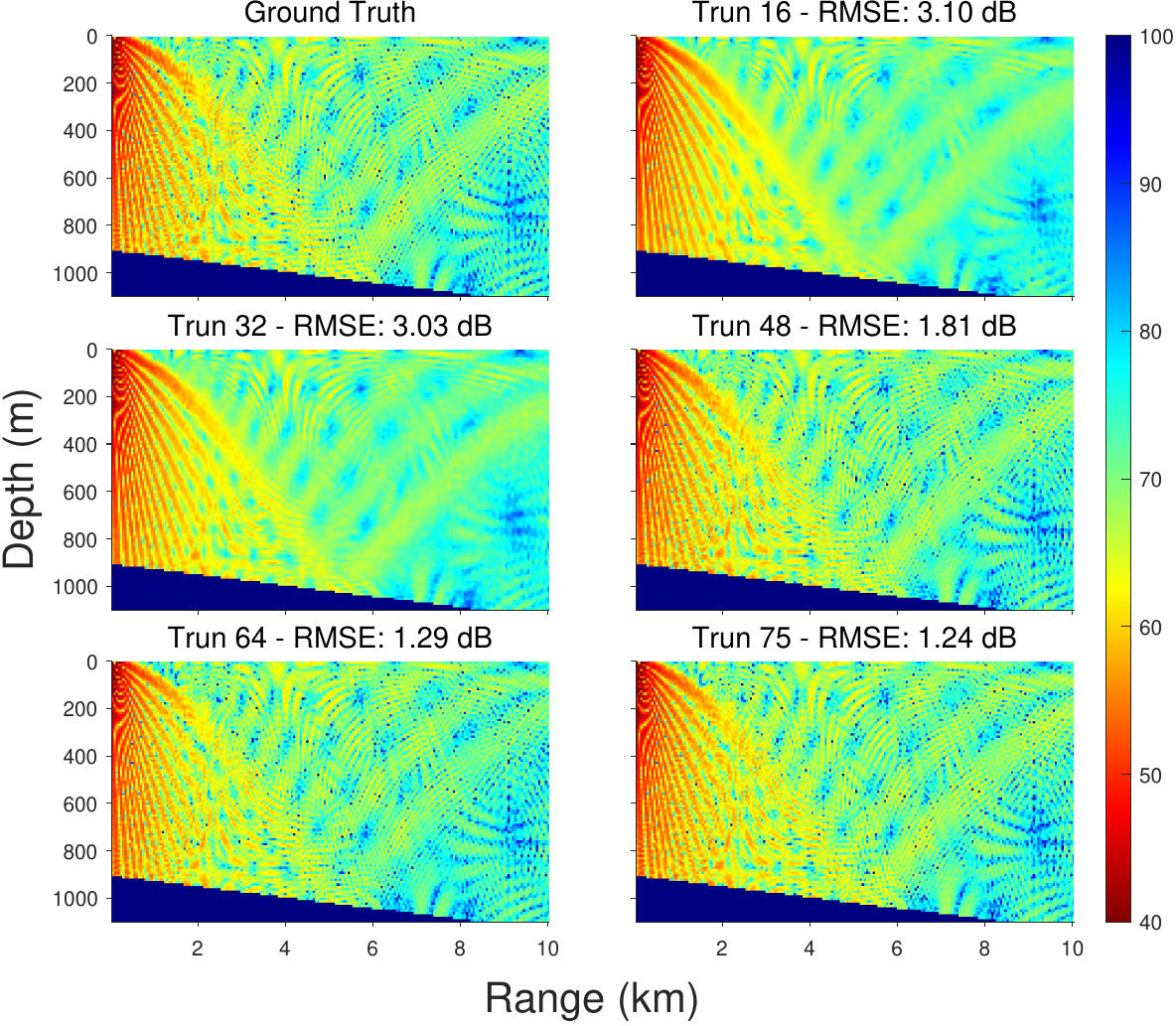}
	\caption{TLs of Hankel-FNO with different truncation mode. "Trun X" denotes truncation up to mode X.}
	\label{fig:ablation_diff_trun}
	\hrule
\end{figure}

\begin{figure*}[t]
	\center
	\includegraphics[width=1.88\reprintcolumnwidth]{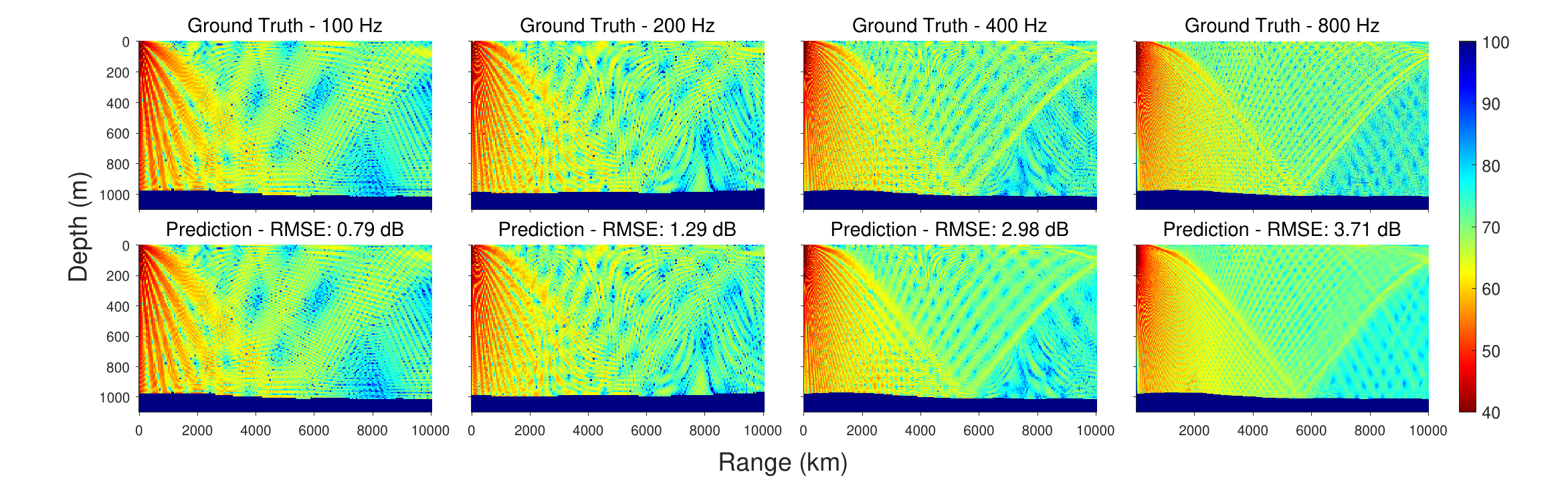}
	\caption{TLs of Hankel-FNO pretrained on datasets with different source frequencies (100, 200, 400, 800 Hz).}
	\label{fig:diff_freq}
	\hrule
\end{figure*}

\begin{figure*}[t]
	\center
	\includegraphics[width=2.0\reprintcolumnwidth]{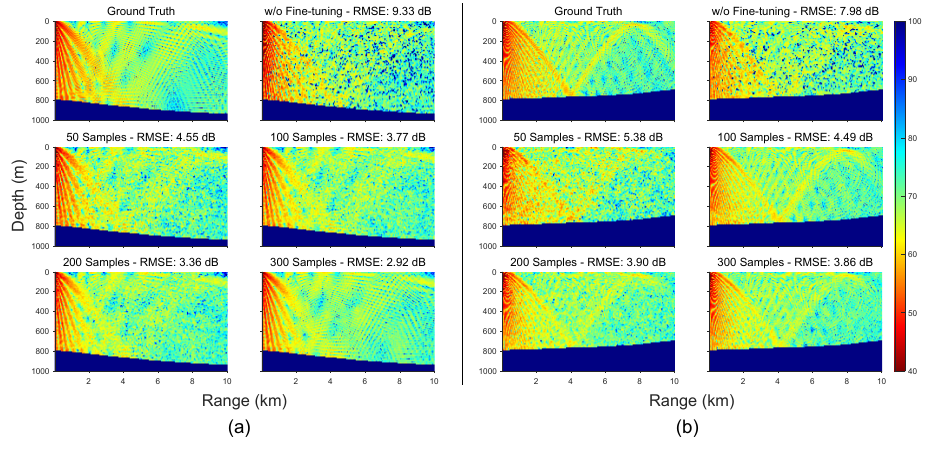}
	\caption{TLs of fine-tuning Hankel-FNO with (a) source frequency 100 Hz and (b) 300 Hz using different amount of samples for 10 km predictions.}
	\label{fig:transfer_freq}
	\hrule
\end{figure*}

\begin{figure*}[t]
	\center
	\includegraphics[width=2.0\reprintcolumnwidth]{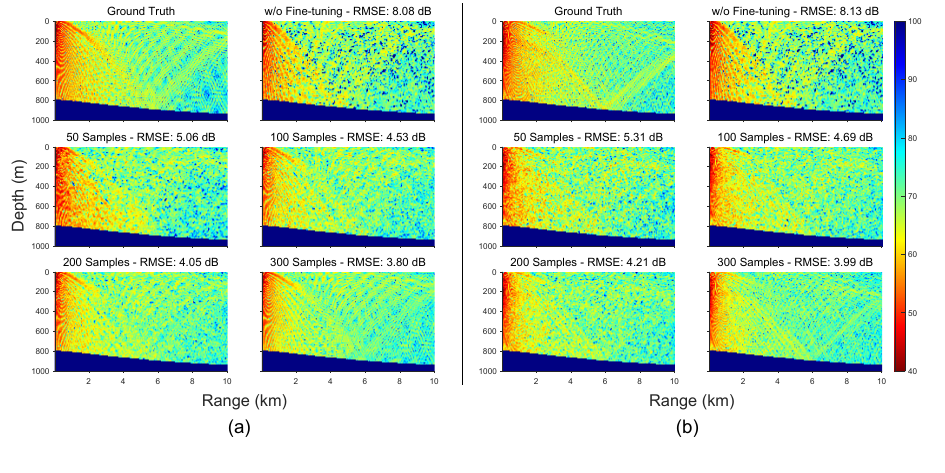}
	\caption{TLs of fine-tuning Hankel-FNO with (a) source depth 100 m and (b) 200 m using different amount of samples for 10 km predictions.}
	\label{fig:transfer_depth}
	\hrule
\end{figure*}

\begin{figure}[t]
	\center
	\includegraphics[width=1.0\reprintcolumnwidth]{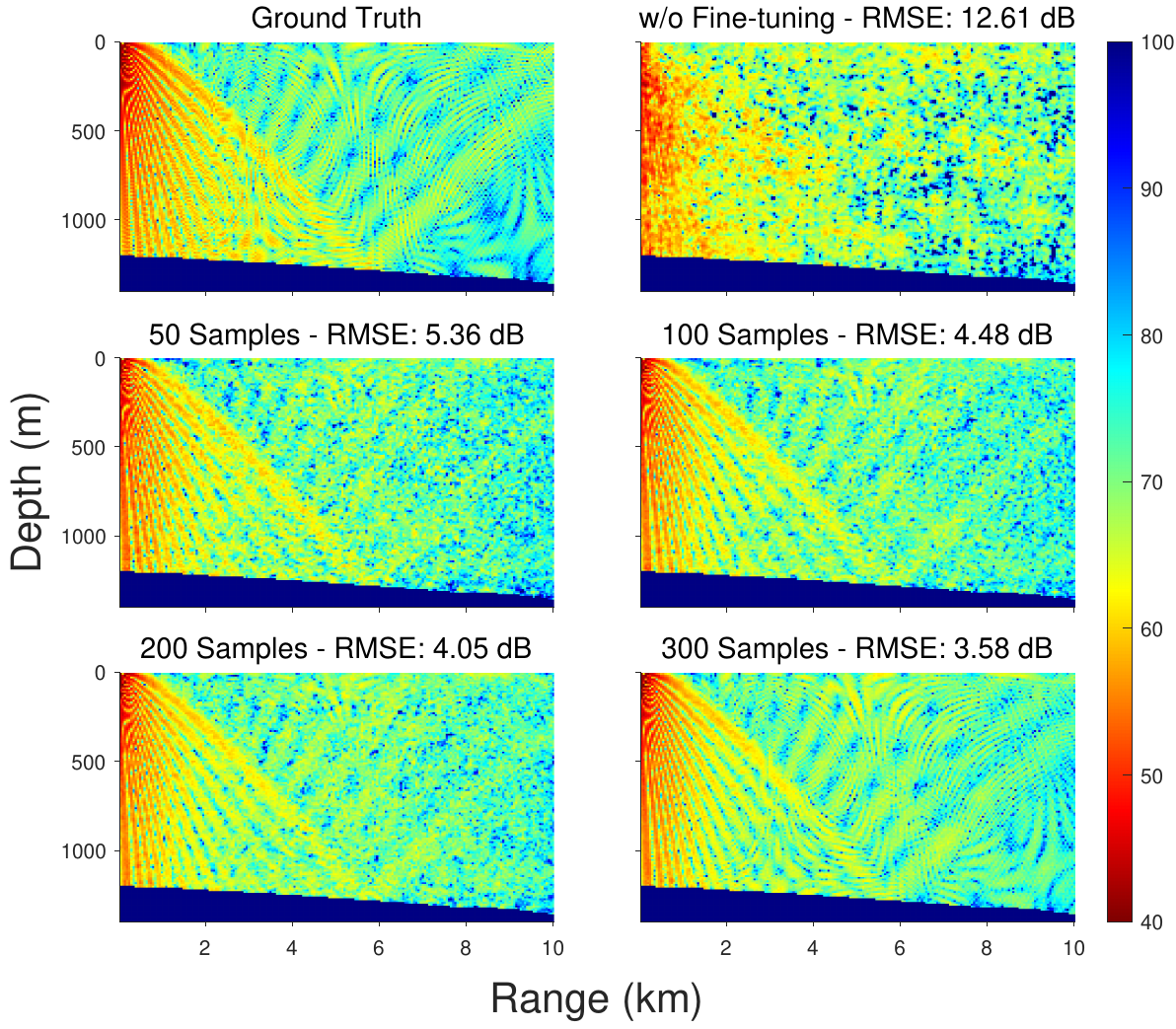}
	\caption{TLs of fine-tuning Hankel-FNO with different bathymetry using different amount of samples for 10 km predictions.}
	\label{fig:transfer_pos5}
	\hrule
\end{figure}

\begin{figure*}[t]
	\center
	\includegraphics[width=2.0\reprintcolumnwidth]{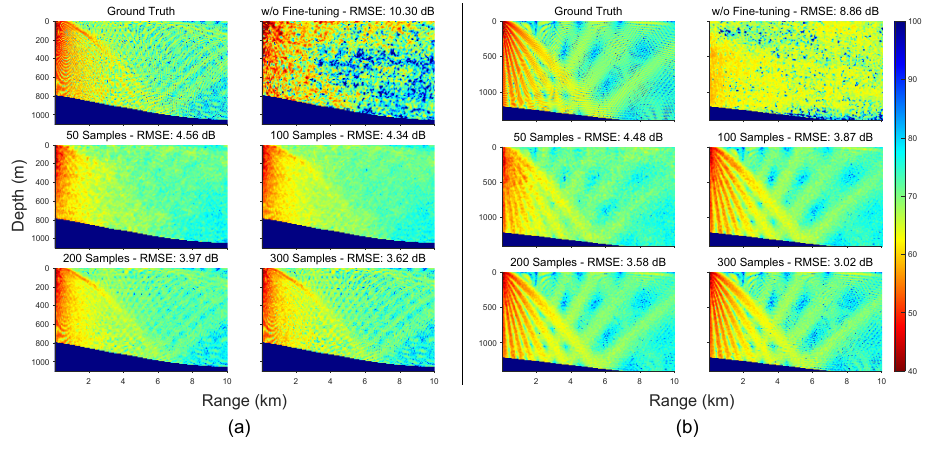}
	\caption{TLs of fine-tuning Hankel-FNO with multiple varying factors: (a) source depth 100 m and source frequency 300 Hz, and (b) different bathymetry and source frequency 100 Hz, using different amounts of samples for 10 km predictions.}
	\label{fig:transfer_hybrid}
	\hrule
\end{figure*}

\begin{figure*}[t]
	\center
	\includegraphics[width=2.1\reprintcolumnwidth]{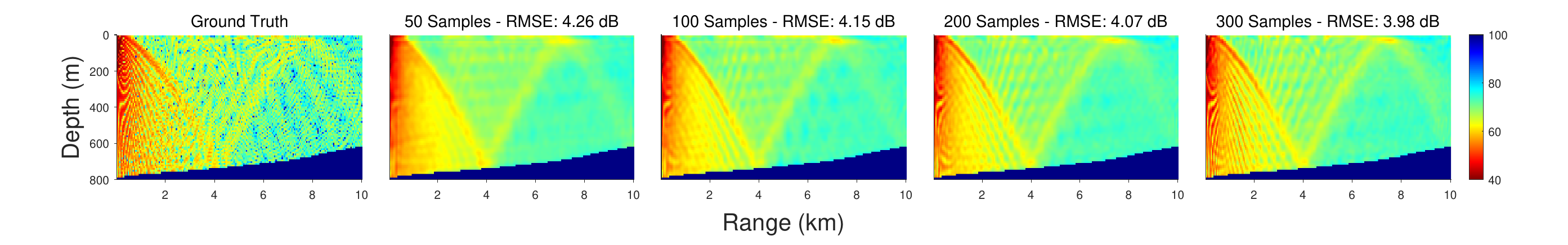}
	\caption{TLs of training Hankel-FNO from scratch on datasets with source frequency 300 Hz using different amount of samples for 10 km predictions.}
	\label{fig:draw_fromscratch}
	\hrule
\end{figure*}

\subsection{\label{sec:4.3} Ablation Study}
\subsubsection{Encoding Strategy Comparison}
To better analyze and demonstrate the contribution of two physics encodings, ablation studies are conducted for different range prediction. 
The performance metrics of FNO with different encodings for 10 km and 40 km TL predictions are in Table~\ref{tab:ablation-10km-40km}, and the generated TLs and error surfaces for 40 km predictions are presented in Fig.~\ref{fig:ablation_visual_effects}. By incorporating bathymetry information, the prediction accuracy for 10 km is significantly improved across all metrics. Environmental information is crucial for modeling sound propagation. Comparing the two bathymetry incorporation strategies, using bathymetry as a separate channel ($C_\text{bty}$) achieves moderate improvement, while bathymetry encoding ($E_\text{bty}$) yields better performance. When bathymetry is treated as an additional channel, the SSF matrix retains redundant information, which hinders feature learning and degrades prediction accuracy. Furthermore, adding an extra channel increases the input dimension, which leads to slower inference speed. These results confirm that integrating bathymetry information directly with SSF data through encoding is more effective.

While the bathymetry encoding also enhances 40 km TL predictions, the improvement remains limited, indicating that additional physical information are required for long-range prediction. This gap is addressed by the Hankel function encoding ($E_\text{hf}$), which provides complementary benefits. While its impact on 10 km TL predictions is moderate, it substantially improves 40 km results, particularly in $H^1$ Sobolev error. The combination of both encodings yields the best performance across all scenarios, demonstrating that environmental and sound propagation encodings are jointly necessary for accurate predictions at varying distances.

\subsubsection{Spectral Truncation Analysis}
The generated TLs and RMSEs of Hankel-FNO with different spectral truncation modes are presented in Fig.~\ref{fig:ablation_diff_trun}. The results demonstrate that the choice of truncation mode significantly affects prediction accuracy. Excessive truncation removes critical spectral information and degrades model performance, while retaining more modes increases computational burden with limited improvement in RMSE.  In our experiments, we select a truncation mode of 64, which achieves a good balance between prediction accuracy and computational efficiency.

\subsection{Pretraining at Different Source Frequencies}
To assess the general applicability of Hankel-FNO, we independently train models at different source frequencies (100 Hz, 200 Hz, 400 Hz, and 800 Hz).
Each model is trained from scratch using the corresponding dataset, without cross-frequency fine-tuning.
We use the 10 km prediction scenario as an example. The experimental settings follow those in Sec.~\ref{sec:4.1.1}. For the 400 Hz and 800 Hz cases, the simulation step size is reduced to ensure numerical stability, and the input-output pair sizes are correspondingly increased to (400, 300) and (1000, 750), respectively.
The results are presented in Fig.~\ref{fig:diff_freq}. Overall, the model demonstrates good predictive performance across all tested frequencies. However, as the source frequency increases, the acoustic field patterns become more complex, significantly increasing the prediction difficulty. At 100 Hz and 200 Hz, the model can accurately predict TL with minimal error. At 400 Hz, some fine-scale details begin to be lost, though the overall structure remains well-preserved. At 800 Hz, where the acoustic patterns are highly complex, detail loss becomes more pronounced, yet the model still successfully captures the main propagation patterns. 

These results reveal a clear frequency-dependent trend.
Consequently, there exists a practical upper limit to the usable frequency, which depends on the specific application requirements: for tasks requiring fine structural details, the model remains reliable below 400 Hz, while for those emphasizing overall propagation patterns, satisfactory results can be achieved up to 800 Hz.
Beyond this range, predictions may become blurred, and architectural adjustments may be necessary.

\subsection{\label{sec:transfer_learning} Fine-tuning}
We demonstrate the rapid transferability of Hankel-FNO across varying acoustic scenarios, including changes in source frequency, source depth, and bathymetry.
In each case, only one factor is varied while the others are held fixed.
The number of training epochs is 100 across all experiments.
To further validate the effectiveness of the two-stage training and fine-tuning strategy, we evaluate its performance against models trained from scratch under the same data and training settings, confirming its adaptability and efficiency.

\subsubsection{\label{sec:source_frequency} Source Frequency}
We consider two scenarios that have a lower (100 Hz) and a higher (300 Hz) frequency than the pretraining (200 Hz). The fine-tuning data is randomly sampled from the training region, with the number of samples varying 50 -- 300 to assess its impact on the accuracy of the fine-tuning process. The number of data used for evaluating the performance is 100 to maintain consistency in the evaluation process. The simulation setup is consistent with the pretraining stage, except for source frequency.
The RMSE of fine-tuning Hankel-FNO with lower or higher source frequencies using different amount of samples for 10 km TL predictions are shown in Table~\ref{tab:finetune_vs_scratch_full} and the generated TLs are in Fig.~\ref{fig:transfer_freq}. The algorithm performs poorly without fine-tuning, because different source frequency leads to different propagation pattern. But the model is able to adapt to a new scenario leveraging fine-tuning with a few training samples and epochs. The results also indicate that the model is better at learning lower-frequency scenario because the variations in sound propagation are smaller.

\begin{table*}[t]
	\centering
	\caption{Quantitative comparison between fine-tuning and training from scratch under various acoustic conditions. 
		Metrics include RMSE, SSIM, and $H^{1}$ Sobolev error $(\times 10^{-2})$ for 10 km TL predictions. The training epoch is set to 100.}
	\label{tab:finetune_vs_scratch_full}
	\renewcommand{\arraystretch}{1.15}
	\setlength{\tabcolsep}{3.5pt}
	\begin{tabular}{c|ccc|ccc|ccc|ccc|ccc}
		\hline\hline
		\multirow{3}{*}{\textbf{Settings}}
		& \multicolumn{6}{c|}{\textbf{Source Frequency}}
		& \multicolumn{6}{c|}{\textbf{Source Depth}}
		& \multicolumn{3}{c}{\textbf{Bathymetry}} \\
		\cline{2-16}
		& \multicolumn{3}{c|}{100 Hz} 
		& \multicolumn{3}{c|}{300 Hz} 
		& \multicolumn{3}{c|}{100 m}
		& \multicolumn{3}{c|}{200 m}
		& \multicolumn{3}{c}{Deeper} \\
		\cline{2-16}
		& RMSE & SSIM & $H^{1}$ 
		& RMSE & SSIM & $H^{1}$ 
		& RMSE & SSIM & $H^{1}$ 
		& RMSE & SSIM & $H^{1}$ 
		& RMSE & SSIM & $H^{1}$ \\
		\hline
		\textbf{w/o fine-tuning} 
		& 9.3 & .55 & 6.8 
		& 8.0 & .51 & 5.7
		& 8.1 & .54 & 5.9 
		& 8.1 & .54 & 5.9 
		& 12 & .37 & 8.3 \\
		\hline
		\textbf{From Scratch (50)} 
		& 3.9 & .67 & 2.7
		& 4.3 & .60 & 2.9 
		& 4.2 & .58 & 3.0
		& 4.2 & .58 & 3.0
		& 4.5 & .60 & 3.0 \\
		\textbf{Fine-tuning (50)} 
		& 4.6 & .71 & 3.3 
		& 5.4 & .70 & 3.3 
		& 5.1 & .65 & 3.7 
		& 5.3 & .63 & 3.9 
		& 5.4 & .64 & 3.7 \\
		\hline
		\textbf{From Scratch (100)} 
		& 3.7 & .69 & 2.6
		& 4.1 & .62 & 2.9 
		& 4.2 & .60 & 3.0 
		& 4.2 & .59 & 3.0 
		& 4.2 & .68 & 2.9 \\
		\textbf{Fine-tuning (100)} 
		& 3.8 & .78 & 2.6 
		& 4.5 & .75 & 2.8 
		& 4.5 & .71 & 3.3 
		& 4.7 & .69 & 3.4
		& 4.5 & .73 & 3.0 \\
		\hline
		\textbf{From Scratch (200)} 
		& 3.6 & .72 & 2.5 
		& 4.1 & .64 & 2.8 
		& 4.1 & .62 & 3.0 
		& 4.2 & .61 & 3.0 
		& 4.0 & .72 & 2.8 \\
		\textbf{Fine-tuning (200)} 
		& 3.4 & .83 & 2.2 
		& 3.9 & .76 & 2.7
		& 4.1 & .75 & 3.0 
		& 4.2 & .73 & 3.1 
		& 4.1 & .78 & 2.5 \\
		\hline
		\textbf{From Scratch (300)} 
		& 3.5 & .76 & 2.5 
		& 4.0 & .67 & 2.8 
		& 4.0 & .66 & 2.9 
		& 4.1 & .62 & 3.0 
		& 4.0 & .75 & 2.7 \\
		\textbf{Fine-tuning (300)} 
		& 2.9 & .87 & 1.9 
		& 3.9 & .81 & 2.3
		& 3.8 & .78 & 2.7 
		& 4.0 & .77 & 2.8 
		& 3.6 & .84 & 2.2 \\
		\hline\hline
	\end{tabular}
\end{table*}

\subsubsection{\label{sec:source_depth} Source Depth}
We consider two scenarios with deeper source depth than the pretraining (50 m): a moderate increase (100 m), and a larger increase (200 m). 
The RMSE of fine-tuning Hankel-FNO with different source depths for different amount of samples are shown in Table~\ref{tab:finetune_vs_scratch_full} and the generated TLs are shown in Fig~\ref{fig:transfer_depth}. Since the model does not take the source depth as an explicit input but learns it implicitly from the data, it fails to adapt to source depth changes. However, with a small amount of fine-tuning data, the model quickly adjusts and generates accurate results.

\subsubsection{\label{sec:bathymetry} Bathymetry}
We consider a scenario where the overall bathymetry is deeper than the pretraining. The fine-tuning data is randomly sampled from another region (purple square in Fig.~\ref{fig:strategy}). The RMSE of fine-tuning Hankel-FNO with different bathymetry for different amount of samples are shown in Table~\ref{tab:finetune_vs_scratch_full} and the generated TLs are shown in Fig.~\ref{fig:transfer_pos5}. The variations in bathymetry affect acoustic wave reflections, leading to distinct propagation patterns. Nevertheless, the model is still able to quickly adapt to scenarios fine-tuning with only a small number of samples.

\subsubsection{\label{sec:hybrid} Multiple Varying Factors}
To further assess the model’s generalization under simultaneous variations of multiple factors, we consider two representative cases: (a) a 100 m source depth combined with 300 Hz source frequency, and (b) deeper bathymetry combined with 100 Hz source frequency. The generated TLs are shown in Fig.~\ref{fig:transfer_hybrid}. Despite the increased complexity introduced by these coupled variations, the model maintains strong predictive performance and achieves high accuracy with fine-tuning on a limited number of samples.

\subsubsection{\label{time_consuming} Fine-Tuning Time}
The average time of fine-tuning Hankel-FNO for different amount of samples are shown in Table \ref{tab:transfer-time}. Although training from scratch is time-consuming, transferring the model to a new scenario is very efficient. 
In addition, more training samples lead to more accuracy TL results, while the fine-tuning time is acceptable.

\subsubsection{\label{scratch_vs_finetune} Fine-Tuning vs. Training from Scratch}
We compare fine-tuned models and models trained from scratch using the same training data and epochs.
As shown in Table~\ref{tab:finetune_vs_scratch_full}, models trained from scratch achieves lower RMSE with small datasets (e.g., 50 training samples) because fine-tuned models require additional adaptation to effectively transfer previously learned features to new scenarios.
However, as the number of samples increases, models trained from scratch encounters a performance bottleneck with limited improvement, while fine-tuned models continues to improve rapidly.
Furthermore, Fig.~\ref{fig:draw_fromscratch} shows the TLs of model trained from scratch with source frequency 300 Hz using different amount of samples.
Due to the neural network's inductive bias, the network initially tends to fit low-frequency information.
Consequently, models trained from scratch only capture the basic sound propagation patterns and fail to model fine details, even with training 300 samples.
This explains why these models perform poorly on SSIM and $H^{1}$ metrics despite achieving relatively low RMSE values.

\begin{table}[t]
	\begin{center}
		\caption{The average time (s) of fine-tuning Hankel-FNO using different amount of samples for 10 km TL predictions.}
		\begin{tabular}{cc}
			\hline\hline
			~Number of Fine-Tuning Data~ & ~Fine-Tuning Time (s)~\\
			\hline
			50 & 48 \\
			\hline
			100 & 87 \\
			\hline
			200 & 160 \\
			\hline
			300 & 244 \\
			\hline\hline
		\end{tabular}
		\label{tab:transfer-time}
	\end{center}
\end{table}

\subsection{\label{sec:4.5} Zero-Shot Inference}
Hankel-FNO can process input data with arbitrary resolution and generating the corresponding output (see Sec.~\ref{sec:3.3}). This enables pretraining on low-resolution data while achieving zero-shot inference at high resolution \cite{li2020fourier}. For example, the environmental data for pretraining have a size of (200, 150), characterizing the ocean region of 10 km range $\times$ 1.5 km depth. To evaluate the zero-shot high-resolution inference capability, we assess unseen environmental data with a size of (200, 300), another ocean region of 10 km range $\times$ 1.5 km depth. The resolution of the test data is twice that of the pretrained data. The higher-resolution input is directly processed by Hankel-FNO.

\begin{figure}[t]
	\center
	\includegraphics[width=0.97\reprintcolumnwidth]{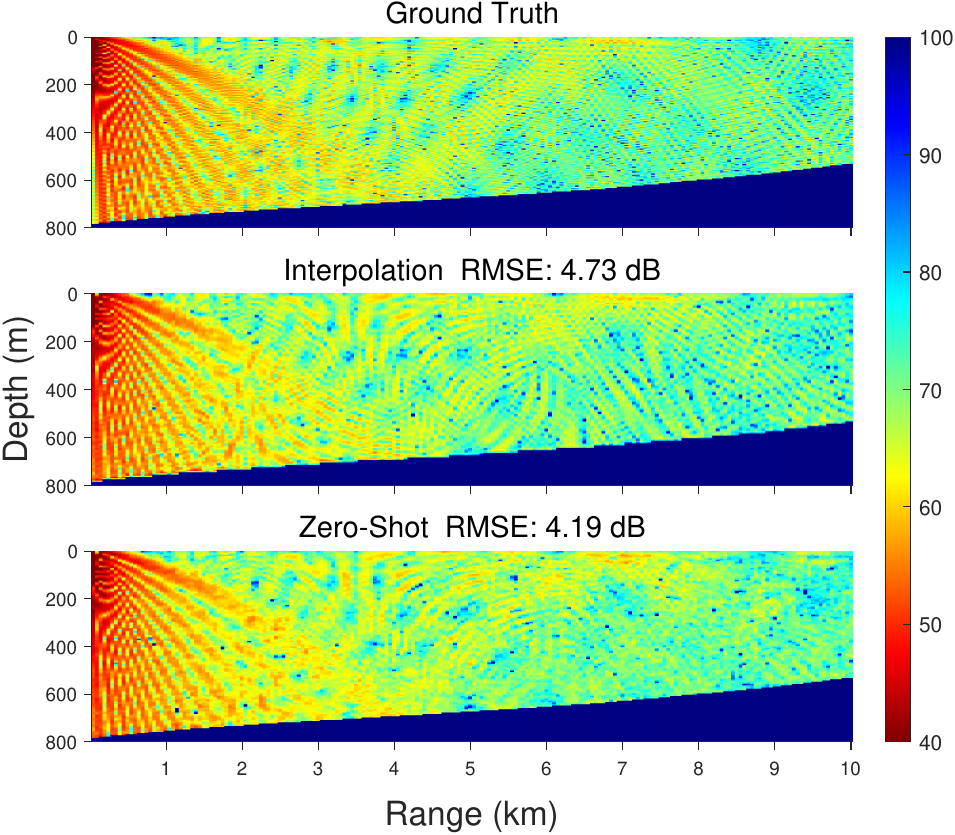}
	\caption{TLs of inferring data with 2$\times$ super-resolution through Bicubic interpolation of low-resolution output and direct processing of high-resolution data using Hankel-FNO.}
	\label{fig:zeroshot}
	\hrule
\end{figure}

A comparison between high-resolution inference using Hankel-FNO and processing via Bicubic interpolation from the low-resolution output is presented in Fig.~\ref{fig:zeroshot}. The low-resolution output is obtained by first downsampling the input environmental field and then applying Hankel-FNO. It is observed that FNO-based method is able to handle data of varying sizes and output accurate results without specialized training. Further, it is clear that the zero-shot high-resolution inference capability of Hankel-FNO is not merely a form of interpolation, but rather a complicated process that capturing the underlying pattern of the data, enabling generalization across resolutions.

\section{\label{sec:5} Conclusion}
We proposed Hankel-FNO, an approach for fast and accurate underwater acoustic charting. By leveraging FNO framework and incorporating key physical encodings -- such as Hankel function and bathymetry information -- Hankel-FNO achieves superior computational efficiency compared to traditional solvers while maintaining higher prediction accuracy than other deep learning-based methods, particularly in long-range scenarios. Furthermore, the model demonstrates generalization, as shown by its ability to adapt to varying environments and source configurations with only a few additional fine-tuning. These advantages make Hankel-FNO a promising solution for real-time and large-scale underwater acoustic charting.

\bibliography{sampbib}

\end{document}